\documentclass[10pt,twocolumn,letterpaper]{article}

\usepackage[pagenumbers]{cvpr}

\usepackage[accsupp]{axessibility}
\usepackage{colortbl}
\DeclareRobustCommand{\modelname}{\textsc{Calico}}
\DeclareRobustCommand{\benchmarkname}{\textsc{MixedParts}}
\usepackage{xspace}
\usepackage{enumitem}
\usepackage{amssymb}
\newcommand{\R}{\mathbb{R}}

\usepackage{booktabs}
\usepackage{tabulary}
\usepackage{multirow}
\usepackage{diagbox}
\usepackage{pgfplots}
\pgfplotsset{compat=1.18}
\usepackage[most]{tcolorbox}
\usepackage{fontawesome5}
\usepackage{listings}

\newcommand{\cc}{\cellcolor{Apricot!50}}
\definecolor{lightgray}{rgb}{0.95, 0.95, 0.95}
\definecolor{darkgray}{rgb}{0.4, 0.4, 0.4}
\definecolor{purple}{rgb}{0.65, 0.12, 0.82}
\definecolor{green}{rgb}{0.0, 0.7, 0.20}
\definecolor{CalicoOrange}{HTML}{F07B41}

\definecolor{object1}{HTML}{5d0b6d}
\definecolor{object2}{HTML}{a8f24a}

\definecolor{common1}{HTML}{c50629}
\definecolor{common2}{HTML}{930f0d}
\definecolor{common3}{HTML}{84eef4}
\definecolor{common4}{HTML}{1804e0}

\definecolor{unique1}{HTML}{84eef4}
\definecolor{unique2}{HTML}{c50629}
\definecolor{unique3}{HTML}{6e0d70}
\definecolor{unique4}{HTML}{3603c9}
\definecolor{unique5}{HTML}{f6fa40}

\definecolor{incontext1}{HTML}{bc0082}
\definecolor{incontext2}{HTML}{a8f24a}

\definecolor{cvprblue}{rgb}{0.21,0.49,0.74}
\usepackage[pagebackref,breaklinks,colorlinks,allcolors=cvprblue]{hyperref}

\def\paperID{17271}
\def\confName{CVPR}
\def\confYear{2025}

\title{\includegraphics[width=1cm]{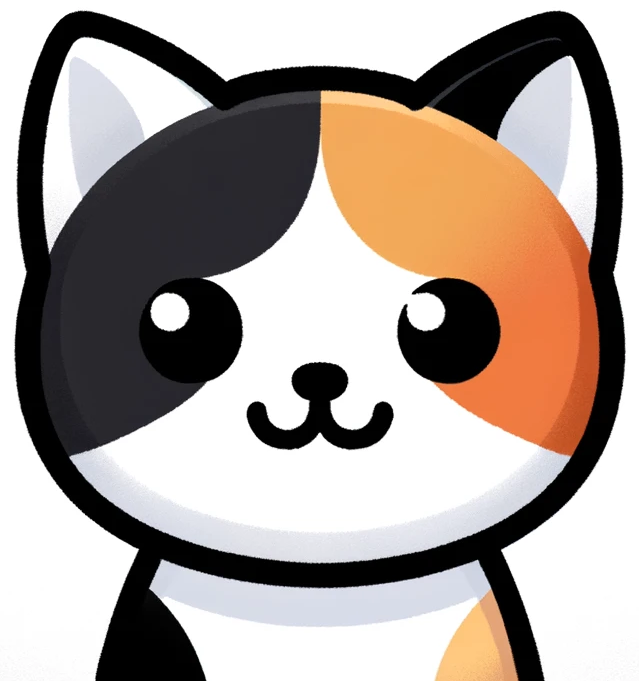} \modelname{}: Part-Focused Semantic Co-Segmentation with ~~~~~~~~~~~~~~~~~~~~~~~~~~~Large Vision-Language Models}
\author{Kiet A. Nguyen\quad
Adheesh Juvekar\quad
Tianjiao Yu\quad
Muntasir Wahed\quad
Ismini Lourentzou \\
\textcolor{CalicoOrange}{University of Illinois Urbana-Champaign} \\
{\tt\small \{kietan2, adheesh2, ty41, mwahed2, lourent2\}@illinois.edu}\\
\small \url{https://plan-lab.github.io/calico}
}

\begin{document}

\twocolumn[{
\renewcommand\twocolumn[1][]{#1}
\maketitle
\begin{center}
    \centering
    \captionsetup{type=figure}
    \includegraphics[width=\linewidth]{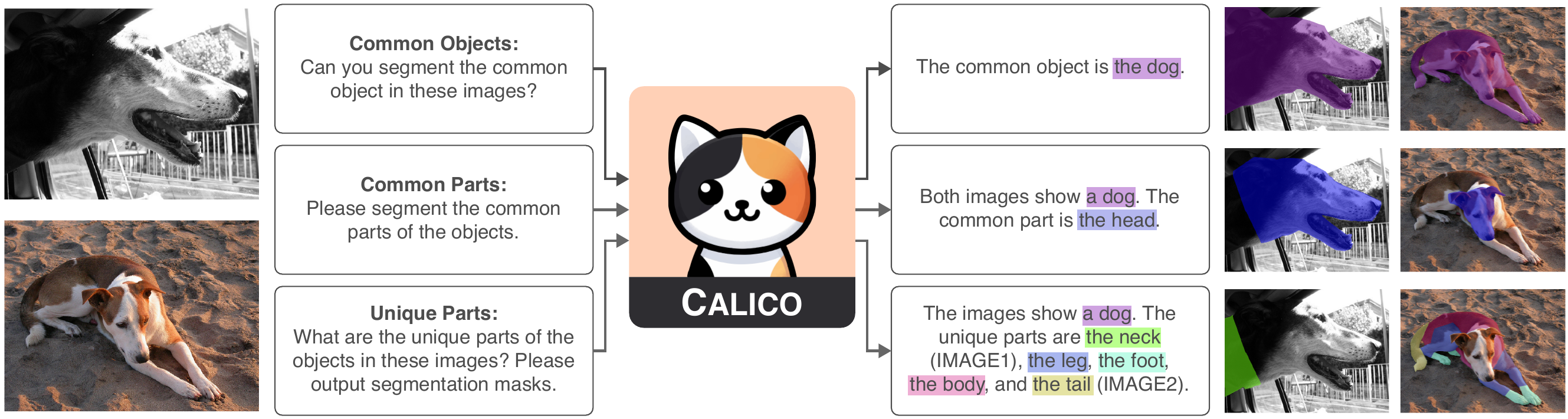}
    \vspace{-0.6cm}
     \captionof{figure}{\textbf{Multi-Image Part-focused Object Comparison with \modelname{}.} Our pixel-grounded Large Vision-Language Model, \modelname{}, performs part-focused semantic co-segmentation, a newly introduced task where the goal is to identify, segment, and label common objects, as well as common and unique object parts across multiple images.}
    \label{fig1}
\vspace{2.5mm}
\end{center}
}]
  
\maketitle

\begin{abstract}
Recent advances in Large Vision-Language Models (LVLMs) have enabled general-purpose vision tasks through visual instruction tuning. 
While existing LVLMs can generate segmentation masks from text prompts for single images, they struggle with segmentation-grounded reasoning across images, especially at finer granularities such as object parts. In this paper, we introduce the new task of \textup{part-focused semantic co-segmentation}, which involves identifying and segmenting common objects, as well as common and unique object parts across images. To address this task, we present \textup{\modelname{}}, the first LVLM designed for multi-image part-level reasoning segmentation. 
\textup{\modelname{}} features two key components, a novel Correspondence Extraction Module that identifies semantic part-level correspondences, and Correspondence Adaptation Modules that embed this information into the LVLM to facilitate multi-image understanding in a parameter-efficient manner. To support training and evaluation, we curate \textup{\benchmarkname{}}, a large-scale multi-image segmentation dataset containing $\sim$2.4M samples across $\sim$44K images spanning diverse object and part categories. Experimental results demonstrate that \textup{\modelname{}}, with just 0.3\% of its parameters finetuned, achieves strong performance on this challenging task.
\end{abstract}    
\section{Introduction}
\label{sec:intro}
Analyzing objects by decomposing them into their constituent parts can enhance understanding of inter-object relationships both within and across categories. This part-level reasoning is crucial for applications requiring detailed object comparisons, such as robotic manipulation, medical imaging, and educational tools. Detailed comparisons that identify shared and unique parts offer insights into the critical features and functions of objects. For example, while both spoons and forks have handles, this part is not central to their primary functions. Instead, distinguishing the fork’s tines from the spoon’s bowl allows tasks like robotic grasping or visual comparisons to differentiate and interact with these objects based on their unique functions.

Designing effective methods to analyze multiple images featuring diverse objects by locating and identifying their shared and distinct parts presents an intriguing challenge. Given a pair of images containing similar objects, the goal is to generate segmentation masks for the shared or distinct parts (\textcolor{MidnightBlue}{locate}), establish one-to-one correspondences across images for comparative analysis (\textcolor{Bittersweet}{compare}), and assign descriptive labels to the discovered parts (\textcolor{Fuchsia}{identify}). We term this task \textbf{\textit{part-focused semantic \underline{co}-segmentation}}.

Existing research has explored various facets of this related granular task. Many works have investigated localized part learning \cite{cho2023partdistillation,van2023pdisconet,li2022panoptic,michieli2022edge,choudhury2021unsupervised,xiao2018unified}, focusing on segmenting parts within an object as a terminal task \cite{cho2023partdistillation,michieli2022edge} or as a means to support broader tasks like object or human parsing \cite{singh2022float,zhou2021differentiable}. 
However, these methods are not designed for multi-image comparisons, as the segmented parts from different images lack a consistent mapping of corresponding semantic components across images. Conversely, some research in this realm has tackled the related task of part co-segmentation, where models learn to segment objects across multiple images in a semantically consistent manner, despite variations in pose, shape, camera angle, \etc \cite{hung2019scops, amir2021deep, liu2021unsupervised, gao2021unsupervised, choudhury2021unsupervised}. Yet, these approaches often require the number of part classes to be specified as input and struggle to identify \textit{unique} parts between objects. 

Recent work has shown a growing interest in leveraging LVLMs for object segmentation tasks \cite{lai2024lisa, rasheed2023glamm, ren2023pixellm, zhang2023next, wei2024lasagna, yuan2023osprey}. Following LISA \cite{lai2024lisa}, many methods introduce additional segmentation tokens \cite{rasheed2023glamm, ren2023pixellm, zhang2023next}, which contain mask embeddings of objects within an image. These tokens are then processed by a decoder to generate the final segmentation masks. Such methods have proven highly effective for unifying tasks like semantic segmentation and referring expression segmentation within a single architecture, thanks to the adaptability of LVLMs in handling diverse input and output types. While some of these models exhibit partial understanding of object parts \cite{lai2024lisa, rasheed2023glamm, yuan2023osprey}, they are not designed for reasoning over parts across multiple images or for segmenting shared and unique parts between objects.

To this end, we introduce \textbf{\underline{C}}omponent-Focused \textbf{\underline{A}}daptive \textbf{\underline{L}}earning for Multi-\textbf{\underline{I}}mage \textbf{\underline{C}}o-Localization of \textbf{\underline{O}}bjects (\textbf{\modelname{})}, a vision-language model designed to perform localized object comparison across image pairs. 
To the best of our knowledge, \modelname{} is the first LVLM trained for multi-image part-level co-segmentation.
\modelname{} augments a pretrained segmentation-based LVLM with parameter-efficient adapter modules to learn multi-image co-localization at multiple granularities. To extract part-level semantic correspondence information between images, we propose a novel Correspondence Extraction Module, which employs a frozen DINOv2 encoder with strong semantic correspondence capabilities learned through self-supervised training \cite{caron2021emerging}. These correspondences are then injected into the pretrained LVLM via Correspondence Adaptation Modules applied at selected layers to facilitate inter-image understanding. 
Due to the lightweight nature of these adaptive modules, the trainable parameters in \modelname{} represent only \textbf{0.3\%} ($\sim$29M) of the entire architecture.

To support training for this novel task, we introduce the \textbf{\underline{M}}ulti-\textbf{\underline{I}}mage \textbf{\underline{Cross}}-S\textbf{\underline{e}}gmentation of \textbf{\underline{D}}istinctive and Common \textbf{\underline{Parts}} (\textbf{\benchmarkname{}}) dataset that contains diverse and logically comparable object pairs that enable \modelname{} to generalize across varied categories and visual details. Building on widely used, publicly available part segmentation datasets \cite{ramanathan2023paco, zhou2017scene, zhou2019semantic, wei2024ov, he2022partimagenet}, we manually curate pairs of object classes that are logically comparable and share at least one common part. For instance, pairing a ``chair'' with an ``ottoman'' is meaningful since both belong to the category of seating furniture, making them more comparable than, \eg, a ``chair'' and a ``microwave.''  We then assemble image pairs based on these class pairings, enabling training for co-segmentation of not only the objects themselves but also their shared and unique parts across multiple images.

\begin{figure}[t!]
  \centering
  \includegraphics[width=0.99\linewidth]{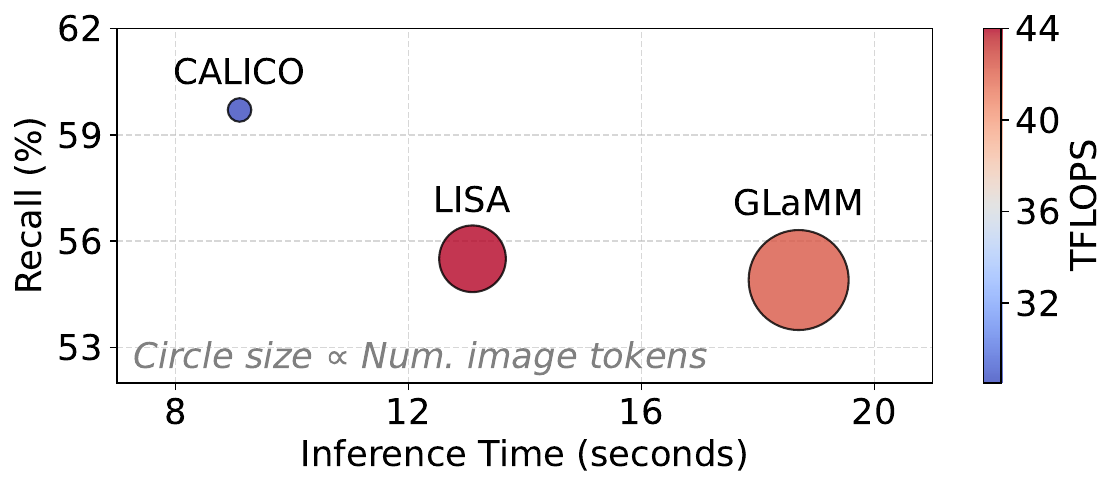}
  \vspace{-0.3cm}
  \caption{\textbf{\modelname{} Efficiency.} \modelname{} improves performance (\eg, recall) in part-focused co-segmentation while reducing TFLOPS by $\sim$32-35\% and accelerating inference by $\sim$30-51\% compared to SotA baselines, using 8-18$\times$ fewer image tokens.}
  \label{fig:tflops}
  \vspace{-0.3cm}
\end{figure}
Since our work is the first to address part-focused semantic co-segmentation, we construct baselines by adapting publicly available pretrained models to this novel task. We conduct extensive experiments on \benchmarkname{} to evaluate the effectiveness of \modelname{}, including ablations to quantify the contributions of the proposed correspondence extraction and adaptation modules. \modelname{} outperforms all baselines on part-focused semantic co-segmentation, achieving a 6.3\% relative improvement in mean IoU on \benchmarkname{} over the next best method, while using 18 times fewer image tokens, reducing TFLOPS by 32.6\%, and accelerating inference by 51.3\% (as shown in Figure \ref{fig:tflops}). In summary, the contributions of our work are as follows:
\begin{itemize}[label=\faPaw, leftmargin=0.65cm, topsep=0cm, partopsep=0pt, parsep=0pt, itemsep=0pt]
    \item We introduce the novel task of \textbf{\textit{part-focused semantic co-segmentation}}, which involves co-segmenting and labeling common and unique parts between objects across images for granular object comparison. To the best of our knowledge, this is the first work to formalize this multi-image object/part co-segmentation task.
    \item We propose \textbf{\modelname{}}, an LVLM for part-focused semantic co-segmentation. \modelname{} incorporates a novel Correspondence Extraction Module and parameter-efficient Adaptation Modules to learn semantic correspondences and enable efficient co-segmentation. 
    \item We curate \textbf{\benchmarkname{}}, a large-scale benchmark for part-focused semantic co-segmentation, featuring logically comparable objects and parts drawn from publicly available diverse part segmentation datasets.
    \item We establish strong baselines from publicly available pretrained models and conduct experiments to evaluate \modelname{}'s performance on \benchmarkname{}, including comprehensive ablations to analyze the contributions of our proposed modules.
\end{itemize}
\section{Related Work}
\label{sec:relatedwork}

\subsection{Prompting Image Segmentation}
Image segmentation, a fundamental computer vision task, has garnered extensive research over the years, with seminal works~\citep{long2015fully,he2017mask,kirillov2019panoptic} paving the way for modern approaches. Pretrained on large-scale data, SAM~\citep{kirillov2023segment} has emerged as a leading foundation model for image segmentation, capable of interpreting diverse prompts and generating corresponding masks. However, despite its strong zero-shot performance, SAM lacks semantic labeling, limiting its practical use. SEEM~\citep{zou2024segment} addresses this by introducing a joint image-text representation space inspired by LVLMs, enabling unified prompting and effective mask labeling. Still, both SAM and SEEM struggle with fine-grained understanding, such as object-part segmentation. Semantic-SAM~\citep{li2023semantic} extends SAM by incorporating multi-level image embeddings tied to different prompt types for multi-granularity understanding but, like SAM, lacks text labels. Recent works~\cite{wei2024lasagna,rasheed2023glamm,lai2024lisa,ren2023pixellm} combine LVLMs with SAM to bridge language and mask understanding, achieving more robust segmentation and labeling capabilities. Building on these advances, \modelname{} integrates SAM with a finetuned pixel decoder to output segmentation masks with text labels. Unlike prior work focused on single-image segmentation, \modelname{} also enables multi-image multi-granularity understanding through object and part co-segmentation.

\subsection{Part Segmentation}
Part segmentation enables fine-grained understanding by decomposing objects into semantically meaningful parts. 
While numerous approaches have been proposed to address this task, many are tailored to specific domains~\citep{yang2020renovating,yang2019parsing,ji2020learning,zhou2023differentiable} or are constrained to closed-set vocabularies~\citep{de2021part,li2022panoptic,michieli2020gmnet,michieli2022edge}. A recent approach, VLPart~\citep{sun2023going}, tackles open vocabulary part segmentation by utilizing a Mask R-CNN backbone~\citep{he2017mask} trained via contrastive learning to align predicted masks with CLIP text features \cite{radford2021learning}, enabling the model to label any object-part. PartGLEE \cite{li2024partglee} adopts a Q-Former-based architecture \cite{li2023blip} to achieve similar alignment of predicted masks with CLIP text features; however, in practice, the labels are constrained by the input to these models' text encoders.
Many recent works~\citep{lai2024lisa, rasheed2023glamm, ren2023pixellm, zhang2023next, wei2024lasagna, yuan2023osprey} integrate the reasoning capabilities of LVLMs with segmentation models to augment visual understanding of the LVLMs. These models enable open-vocabulary object segmentation without being constrained to the input prompt. Most of these models, motivated by LISA~\citep{lai2024lisa}, utilize LLaVA-based LVLMs ~\citep{liu2024visual} in tandem with a decoder to predict segmentation masks.

While several of these works can also segment object parts~\citep{lai2024lisa,rasheed2023glamm,yuan2023osprey} -- since their training data consists of large annotated datasets with part information -- their part segmentation capabilities often lag behind their object segmentation capabilities. Additionally, they also require explicit mention of object parts in the input prompt for successful segmentation and fail to segment multiple parts of an object. Instead, each part must be individually specified, \eg, by requesting to ``segment the leg of the chair,'' to obtain the desired segmentation maps. In contrast, \modelname{} augments a pretrained LVLM with object and part co-segmentation capabilities without the need for explicit part-specific prompts, simplifying user input requirements and allowing flexibility in instructing the model to infer and delineate various objects and parts across different images.

\subsection{Object/Part Co-Segmentation}
\begin{figure*}[t!]
  \centering
  \includegraphics[width=\linewidth]{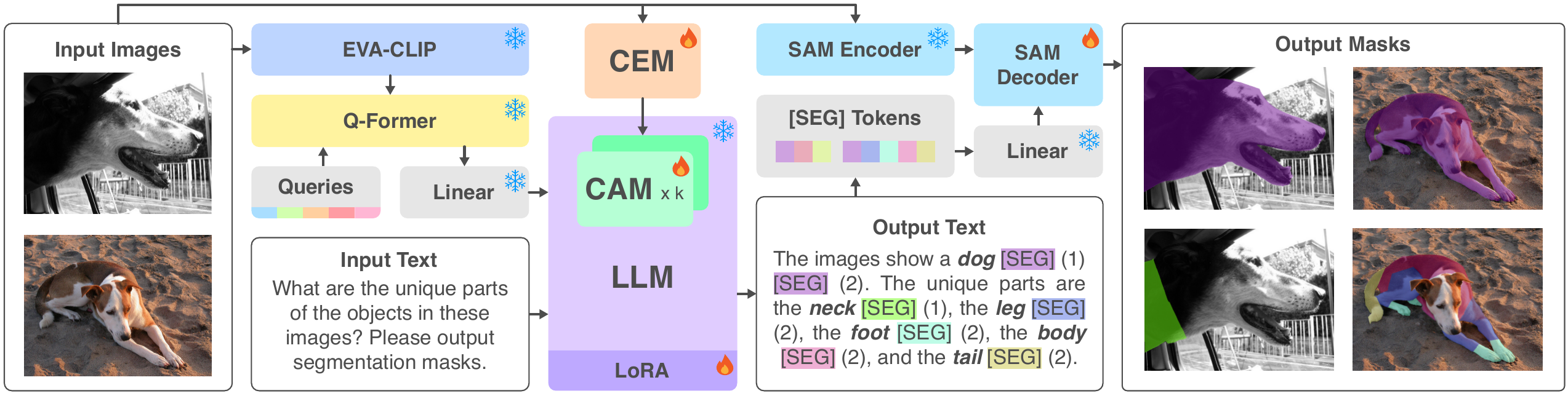}
  \vspace{-0.6cm}
  \caption{\textbf{Overview of the \modelname{} Architecture for Part-Focused Semantic Co-Segmentation.} \modelname{} employs a Q-Former cross-attention module to query efficient image embeddings from a pretrained image encoder, which are passed as visual tokens into a Vicuna-based LLM. We extract \texttt{[SEG]} tokens from the output text, which are used to prompt a SAM decoder to produce corresponding segmentation masks. We propose two modules: the Correspondence Extraction Module (CEM), which captures semantic-rich part correspondences, and Correspondence Adaptation Modules (CAMs), which inject this information into the LVLM. CEM/CAM details in Figure \ref{fig:cemcam}.}
  \label{fig:arch-overview}
  \vspace{-0.3cm}
\end{figure*}
Co-segmentation aims to identify and segment common objects across multiple images. Early works~\citep{li2019group,zhang2020deep,zhang2021cyclesegnet} employ CNNs with fully supervised training or finetuning on co-segmentation datasets, while later works leverage the semantic understanding of CLIP~\citep{duan2023lcco} or finetune an LVLM on samples containing similar objects \cite{pramanick2024jack}. However, these methods do not extend to co-segmenting object parts. 
Conversely, part co-segmentation involves simultaneously segmenting corresponding object parts across multiple images. Previous methods tackle this task with unsupervised or self-supervised learning~\citep{hung2019scops, amir2021deep, liu2021unsupervised, gao2021unsupervised, choudhury2021unsupervised}. DFF~\citep{collins2018deep} employs matrix factorization on features from a pretrained CNN, while SCOPS~\citep{hung2019scops} trains an encoder-decoder CNN with equivariance and semantic consistency losses for improved part co-segmentation. More recently, \citep{amir2021deep} shows that DINO features capture semantic and spatial correspondences useful for identifying similar parts across images. 

While effective at discovering part-level correspondences, these methods lack the ability to assign semantic part labels or generate segmentation masks for unique parts.
In this work, we introduce \modelname{}, a model that harnesses SAM's segmentation capabilities and DINOv2's semantic features \cite{oquab2023dinov2, darcet2023vision} to jointly perform part co-segmentation and labeling. By integrating LVLMs with co-segmentation, \modelname{} improves object-part reasoning across images with diverse scenes, generating both co-segmentation masks and semantic labels for unique and common parts.
\section{Method}
\subsection{Problem Definition}
In part-focused semantic co-segmentation, the objective is to generate a set of segmentation masks for a given set of input images. Here, each mask corresponds to an input image and indicates the pixels containing either the common object across all images or the shared and distinct parts belonging to semantically similar objects (\ie, intuitively comparable objects) within these images. Formally, given $N_I$ input images $\mathbf{X}_\textnormal{image}\!=\!\left\{\mathbf{X}_\textnormal{image 1}, \cdots, \mathbf{X}_{\textnormal{image }N_I}\right\}$, where each $\mathbf{X}_{\textnormal{image }i} \in \R^{3 \times H_i \times W_i}$ has height $H_i$ and width $W_i$, the goal is to train a co-segmentation model $\mathcal{F}\!:\!\mathbf{X}_\textnormal{image} \mapsto \mathbf{M}$ to obtain a set of mask sets $\mathbf{M}\!=\!\{\mathbf{M}_1, \cdots, \mathbf{M}_{N_I}\}$, with each $\mathbf{M}_i\!=\!\left\{\left(\mathbf{m}_{i1}, c_{i1}\right), \cdots, \left(\mathbf{m}_{iM_j}, c_{iM_j}\right)\right\}$ containing $M_j$ masks and corresponding class labels associated with image $i$. Here, each binary mask $\mathbf{m}_{ik} \in \{0,1\}^{H_i \times W_i}$  assigns each pixel to a value of $1$ if it covers the visual element with semantic class label $c_{ik}$ and $0$ otherwise. We denote $\mathbf{c}_{i}\!=\!\left\{c_{i1}, \cdots, c_{iM_j}\right\}$ as the set of class labels corresponding to image $i$.
When generating common object or part masks across all images, we want $\bigcap_{i=1}^{M_j} \mathbf{c}_{i}\!\neq\!\varnothing$, whereas when obtaining unique masks, we want $\mathbf{c}_{i} \cap \mathbf{c}_{i'}\!=\!\varnothing\; \forall i\!\neq\!i', 1\!\leq\!i, i'\!\leq\!M_j$. To learn a model $\mathcal{F}$ that can address this multifaceted task, we opt for an LVLM-based solution, leveraging LLMs' ability to tackle multiple tasks with a single architecture and their flexibility in input/output processing. The rest of this section details our model's architecture.

\subsection{\textbf{\modelname{}} Architecture}
\label{sec:model-arch}
\modelname{} is an LVLM designed to generate multiple segmentation masks per image across a series of images, highlighting both shared and distinct regions among them. In addition to its core functionality, our model incorporates modules aimed at integrating semantic correspondences between similar objects across images, alongside multi-image understanding and segmentation. 
As illustrated in Figure~\ref{fig:arch-overview}, the architecture is composed of a Vicuna-based LLM $\mathcal{M}$ in tandem with a vision module $\mathcal{I}$ and a vision-to-language projection layer, which projects image embeddings from $\mathcal{I}$ into $\mathcal{M}$'s language space.

\noindent \textbf{Interleaved Vision-Language Inputs.} \modelname{} is trained to understand interleaved multi-image inputs. Given $N_I$ input images $\mathbf{X}_\textnormal{image} \in \R^{N_I \times 3 \times H \times W}$ and a vision module $\mathcal{I}\!:\!\R^{3 \times H \times W} \mapsto \R^{S_I \times D_I}$, we obtain image embeddings $\mathbf{X}_\textnormal{embed} \in \R^{N_I \times S_I \times D_I}$ by $\mathbf{X}_\textnormal{embed}\!=\!\mathcal{I}\left(\mathbf{X}_\textnormal{image}\right)$, where $S_I$ and $D_I$ are the image embedding sequence length and hidden size, respectively. We then project these embeddings into the language model space with hidden size $D$ via $f_{\textnormal{image}}\!:\!\R^{D_I} \mapsto \R^{D}$ to get
\begin{equation}
\scalebox{0.95}{$
    \mathbf{I}^0\!=\!f_{\textnormal{image}}\left(\mathbf{X}_\textnormal{embed}\right)\!=\!\left\{ \mathbf{I}^0_1, \cdots, \mathbf{I}^0_{N_I} \right\} \in \R^{N_I \times S_I \times D}.
     $}
\end{equation}

The final input $\mathbf{T}^0\!\in\!\R^{S \times D}$ into $\mathcal{M}$ is composed of interleaved text and image tokens 
$\mathbf{T}^0\!=\!\{\mathbf{t}^0_1, \cdots, \mathbf{v}^0_{11},$ $\cdots, \mathbf{v}^0_{1S_I}, \cdots,$ $ \mathbf{t}^0_i, \cdots, \mathbf{v}^0_{j1}, \cdots, \mathbf{v}^0_{jS_I}, \cdots, \mathbf{t}^0_{S_T}\}$,
where $\mathbf{t}^0_i$ is the $i^{\textnormal{th}}$ embedded text token at layer $0$ ($1 \leq i \leq S_T$), $\mathbf{I}^0_j\!=\!\left\{\mathbf{v}^0_{j1}, \cdots, \mathbf{v}^0_{jS_I}\right\}$ are the $S_I$ tokens pertaining to the $j^{\textnormal{th}}$ image, and the superscript $k$ of $\mathbf{T}^k$ represents the LLM output at layer $k$ or input at layer $k + 1$. The sequence length $S$ of $\mathbf{T}^0$ thus sums up the text and image lengths, \ie, $S\!=\!S_T + N_I \times S_I$. Finally, we obtain predicted outputs from the $N$-layered LLM by $\mathbf{\hat{T}}^N\!=\!\mathcal{M}\left(\mathbf{T}^0\right)$.
In practice, we encourage the LLM to develop comprehensive multimodal understanding through prompts such as
{``\texttt{The <image> (IMAGE1) and <image> (IMAGE2) provide an overview of the pictures. Can you segment the common object in these images?}'',}
where the \texttt{<image>} tokens are replaced with the projected embeddings of the corresponding image. Each image is also associated with a unique identifier (\eg, \texttt{IMAGE1}, \texttt{IMAGE2}) for more convenient and clear reference, avoiding any potential ambiguity from using ordinal terms to refer to individual images in multi-image settings, \eg, ``the first'' or ``the second'' image. 

\noindent \textbf{Vision Module.} 
Observing the effectiveness and efficiency of BLIP-2's Q-Former cross-attention mechanism \cite{li2023blip}, especially in multi-image settings \cite{li2023fine}, we propose using Q-Former in tandem with a strong CLIP vision encoder \cite{sun2023eva, radford2021learning} to extract visual embeddings from the input images. Whereas projecting CLIP embeddings directly into the language model space preserves their long sequence lengths (\eg, 256 or 576 tokens \cite{lai2024lisa, rasheed2023glamm}) and thus increasing compute, Q-Former uses a much shorter set of learnable query tokens to extract visual information (\eg, 32 tokens \cite{li2023blip, li2023fine}).
Formally, our vision module $\mathcal{I}$ consists of an EVA-CLIP-g model $\mathcal{C}\!:\!\R^{3 \times H \times W} \mapsto \R^{S_C \times D_C}$ and a Q-Former cross-attention module $\mathcal{Q}$ alongside a set of learnable query tokens $\mathbf{q} \in \R^{S_I \times D_I}$. We first pass input images $\mathbf{X}_\text{image}$ through the EVA-CLIP global encoder to obtain $\mathbf{X}_\text{global}\!=\!\mathcal{C}\left(\mathbf{X}_\text{image}\right) \in \R^{N_I \times S_C \times D_C}$. We then obtain our final visual embeddings by querying $\mathcal{Q}$ using $\mathbf{q}$ as the query and $\mathbf{X}_\text{global}$ as the key and value:
\begin{equation}
    \mathbf{X}_\text{embed}\!=\!\mathcal{Q}\left(\mathbf{q}, \mathbf{X}_\text{global}\right) \in \R^{N_I \times S_I \times D_I}.
\end{equation}
\noindent \textbf{Pixel-Grounded Outputs.}
To enable pixel-level grounding, we augment the model's vocabulary with the segmentation token \texttt{[SEG]} and teach it to output grounding tags \texttt{<p>} and \texttt{</p>}, following recent work \cite{rasheed2023glamm}. Through supervision, the model learns to ground a noun phrase associated with the following segmentation token by enclosing it in the grounding tags, which immediately precede the corresponding \texttt{[SEG]} token (\eg, ``\texttt{The unique parts of the objects are <p> the seat cushion </p> [SEG] (IMAGE1) and <p> the back pillow </p> [SEG] (IMAGE2).}''). We append the image identifiers immediately following \texttt{[SEG]} tokens to distinguish between tokens belonging to different images.
To transform the \texttt{[SEG]} tokens $\mathbf{S}\!=\!\left\{\mathbf{S}_1,\!\cdots,\!\mathbf{S}_{N_I}\right\} \subset \mathbf{T}^K$, where each $\mathbf{S}_i \in \R^{S_j \times D}$ corresponds to the set of $S_j$ predicted masks associated with image $i$, into segmentation mask sets $\mathbf{\hat{M}}\!=\!\left\{\mathbf{\hat{M}}_1, \cdots, \mathbf{\hat{M}}_{N_I}\right\}$, our architecture incorporates a Transformer-based grounding model \cite{kirillov2023segment}, composed of a grounding encoder $\mathcal{G}$ and a pixel decoder $\mathcal{D}$. The input images $\mathbf{X}_\textnormal{image}$ are passed through the frozen encoder to obtain grounding embeddings $\mathbf{X}_\textnormal{ground}$ as the vision signal for the decoder by $\mathbf{X}_\textnormal{ground}\!=\!\mathcal{G}\left(\mathbf{X}_\textnormal{image}\right)\!\in\!\R^{N_I \times S_D \times D_D}$. For an encoded image $\mathbf{X}_{\text{ground } i}$, the tokens in $\mathbf{S}_i$ act as prompts for the finetuned pixel decoder after being passed through a projection layer $f_\text{segmentation}\!:\!\R^D \mapsto \R^{D_D}$. Finally, the decoder $\mathcal{D}$ produces binary segmentation masks accordingly by:
\begin{equation}
    \mathbf{\hat{M}}_i = \mathcal{D}\left(\mathbf{X}_{\textnormal{ground } i}, f_\text{segmentation}(\mathbf{S}_i)\right).
\end{equation}

\subsection{Correspondence Extraction Module (CEM)}
\label{sec:correspondence-extraction}
\begin{figure}[t!]
  \centering
  \includegraphics[width=0.75\columnwidth]{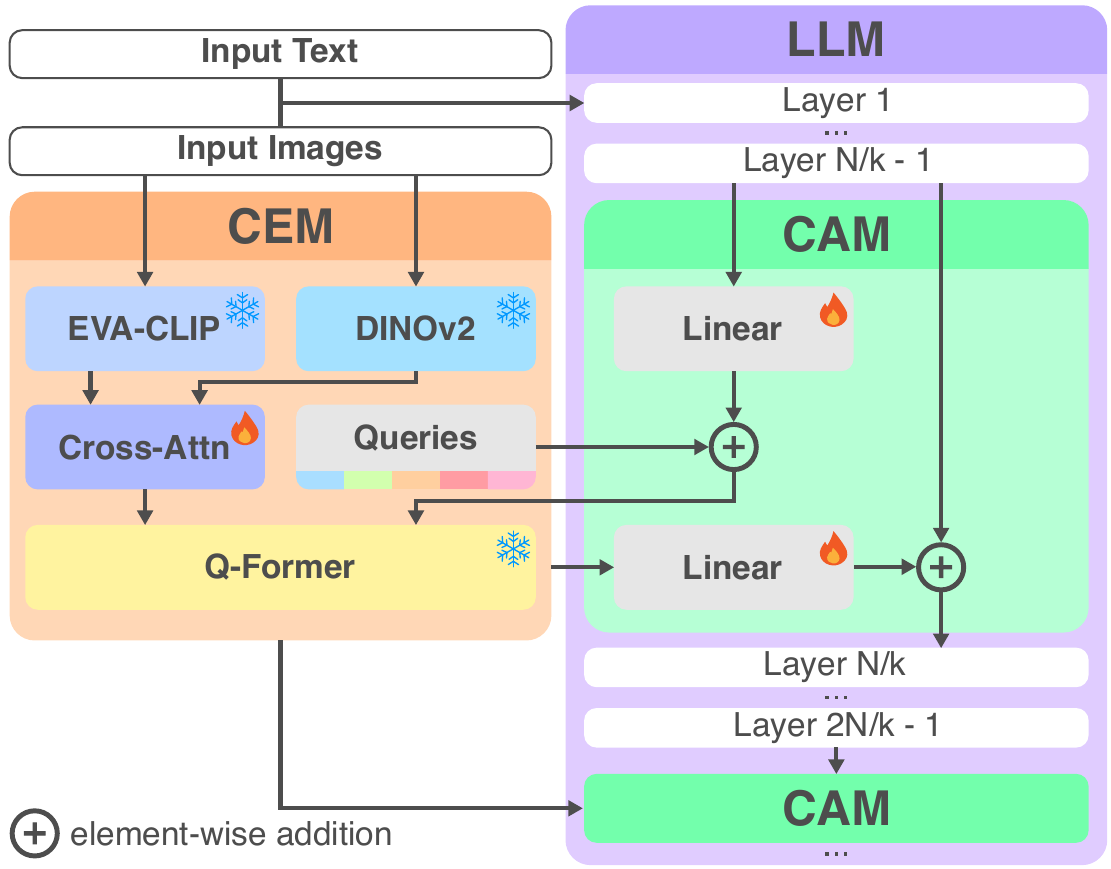}
  \vspace{-0.2cm}
  \caption{\textbf{Overview of our Correspondence Extraction and Correspondence Adaptation Modules.} In \modelname{}, $k$ CAMs are placed at every $\frac{N}{k}$ layers in the $N$-layered LLM.}
  \label{fig:cemcam}
  \vspace{-0.3cm}
\end{figure}

Image features obtained from self-supervised Vision Transformers (ViTs) \cite{caron2021emerging, oquab2023dinov2, darcet2023vision} have been shown to exhibit rich semantic information at part-level granularity across similar, yet distinct object categories \cite{amir2021deep, caron2021emerging, zhang2024tale}. Motivated by these findings, we design a fusion module to extract such semantic information and facilitate part correspondence learning within the model. We define $\mathcal{E}$ to be the semantic extraction process using a self-supervised ViT \cite{caron2021emerging} similarly to \cite{amir2021deep}, and obtain semantic embeddings $\mathbf{X}_\textnormal{semantic} \in \R^{N_I \times S_S \times D_S}$, where $S_S$ and $D_S$ are the sequence length and hidden size of the semantic image embeddings, respectively. We then use $\mathbf{X}_\textnormal{semantic}$ as the key and value for a cross-attention extraction mechanism $\mathcal{A}$ with the queried EVA-CLIP embedding $\mathbf{X}_\text{global}$ to get semantic-rich global embeddings $\mathbf{X}'_\text{global}$. This process is formalized by:
\begin{align}
    \mathbf{X}_\text{global} & = \mathcal{C}\left(\mathbf{X}_\text{image}\right) & \in \R^{N_I \times S_C \times D_C} \\
    \mathbf{X}_\text{semantic} & = \mathcal{E}\left(\mathbf{X}_\text{image}\right) &\in \R^{N_I \times S_S \times D_S} \\
    \mathbf{X}'_\text{global} & = \mathcal{A}\left(\mathbf{X}_\text{global}, \mathbf{X}_\text{semantic}\right) & \in \R^{N_I \times S_C \times D_C}
\end{align}
\noindent This fusion process produces strong semantic embeddings $\mathbf{X}'_\text{global}$, which are subsequently utilized for the visual extraction process performed by the Correspondence Adaptation Modules, detailed in the next section.

\begin{table*}[t!]
\centering
    \resizebox{\textwidth}{!}{
\begin{tabular}{>{\centering\arraybackslash}m{0.04\textwidth}|>{\centering\arraybackslash}m{0.4\textwidth}|>{\centering\arraybackslash}m{0.4\textwidth}|>{\centering\arraybackslash}m{0.4\textwidth}}
  & \multicolumn{1}{c|}{\textbf{PACO-LVIS}} & \multicolumn{1}{c|}{\textbf{PartImageNet}} & \multicolumn{1}{c}{\textbf{ADE20K-Part-234}} \\
  \multirow{3}{*}{\rotatebox{90}{\parbox{0.1\textwidth}{\centering\textbf{Object}}}} &
  \begin{minipage}[c][6em][c]{\linewidth}
    \centering
    \includegraphics[height=6em, width=0.5\linewidth]{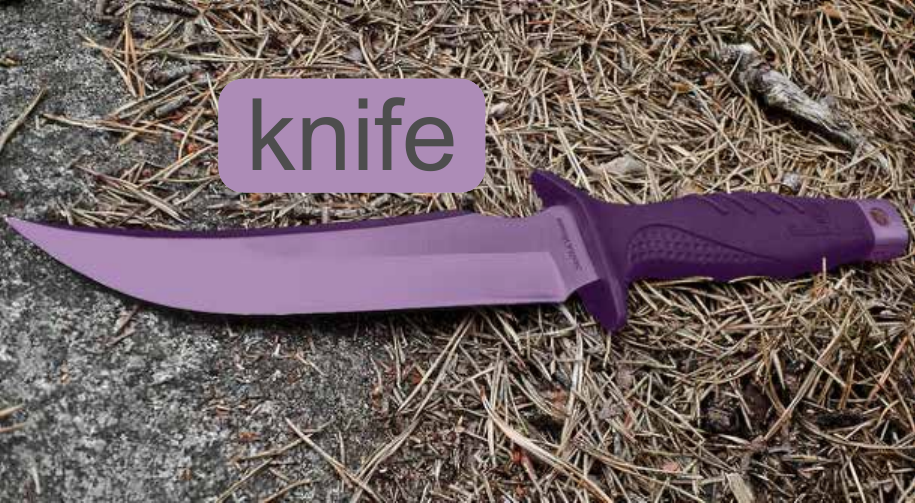}%
    \includegraphics[height=6em, width=0.5\linewidth]{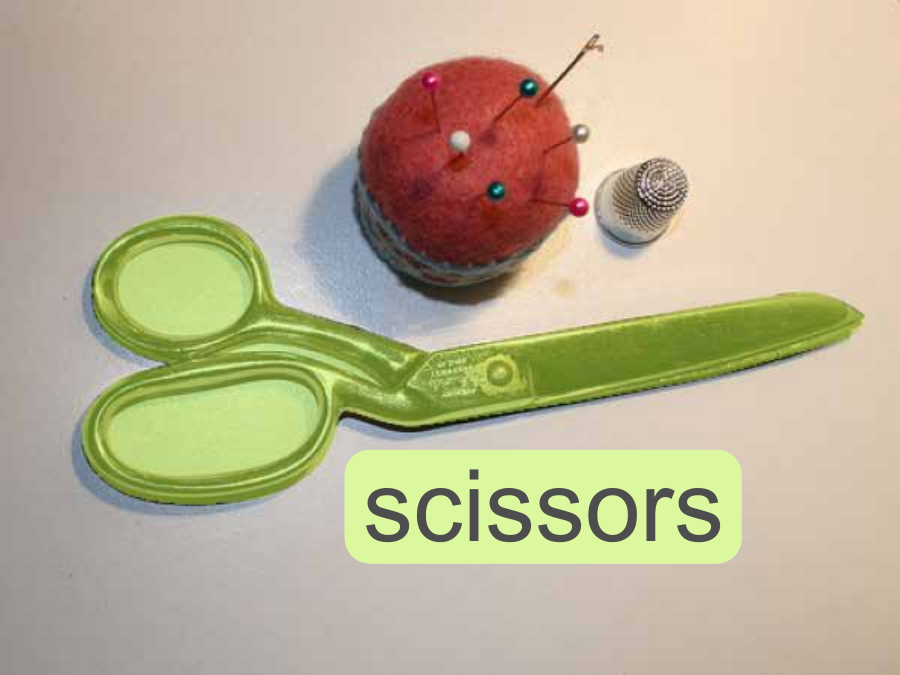}
  \end{minipage} &
  \begin{minipage}[c][6em][c]{\linewidth}
    \centering
    \includegraphics[height=6em, width=0.5\linewidth]{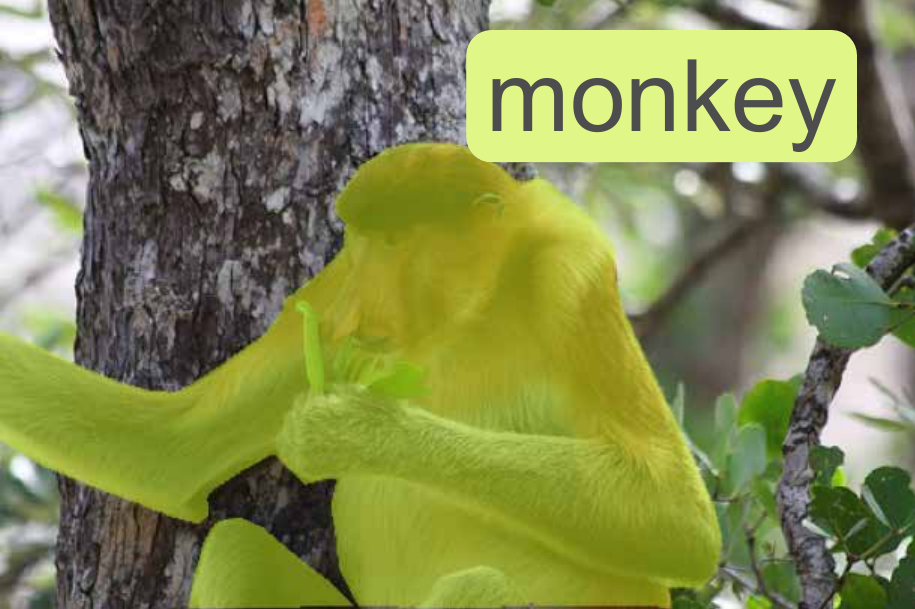}%
    \includegraphics[height=6em, width=0.5\linewidth]{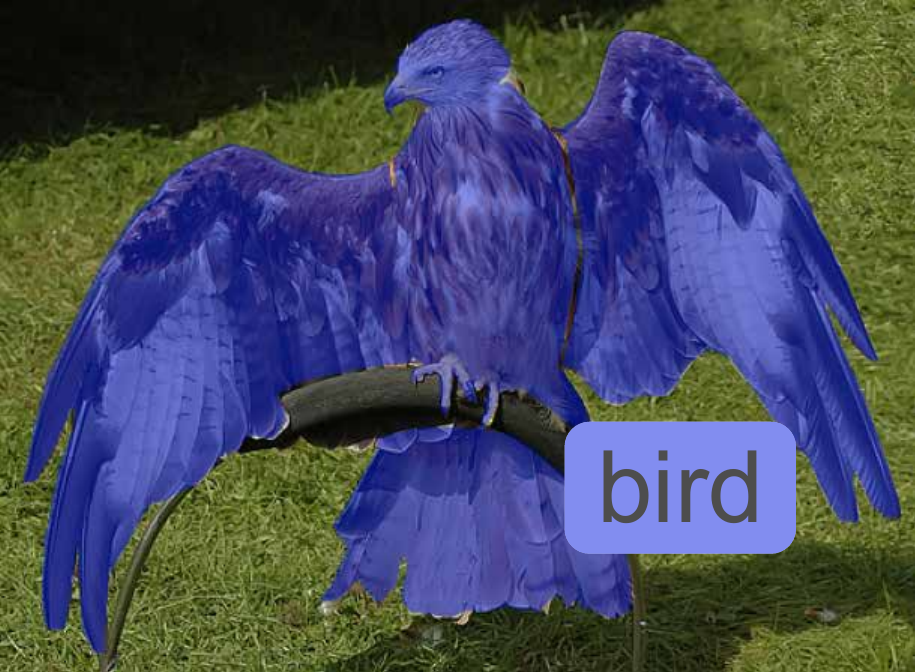}
  \end{minipage} &
  \begin{minipage}[c][6em][c]{\linewidth}
    \centering
    \includegraphics[height=6em, width=0.5\linewidth]{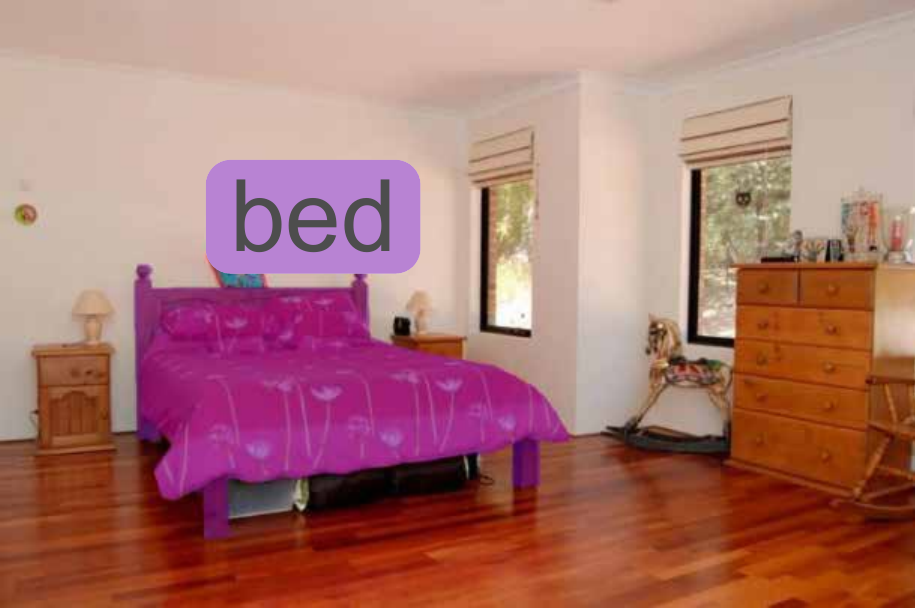}%
    \includegraphics[height=6em, width=0.5\linewidth]{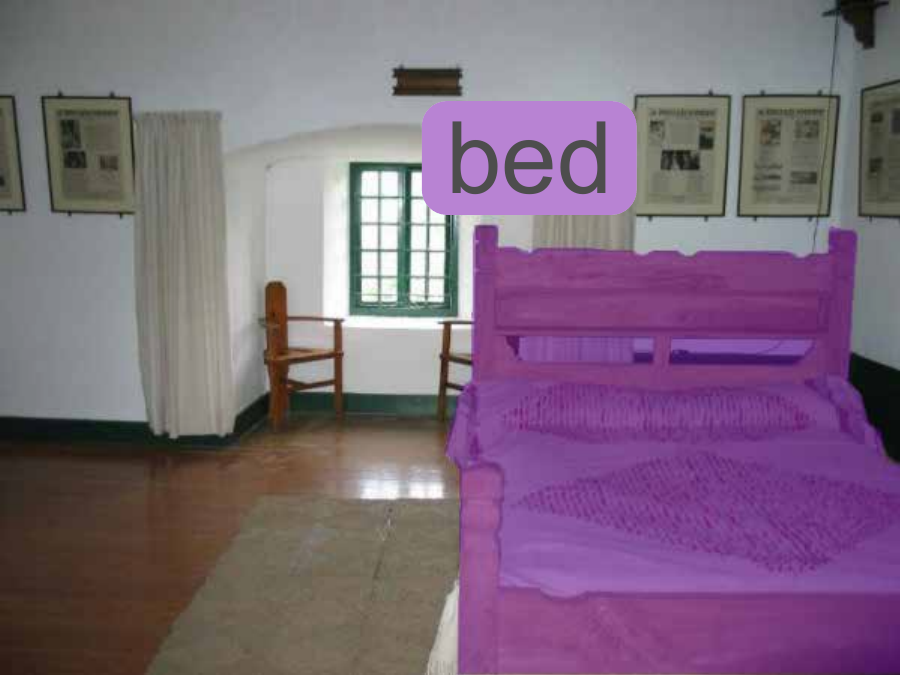}
  \end{minipage} \\
  \multirow{3}{*}{\rotatebox{90}{\parbox{0.1\textwidth}{\centering\textbf{Common Parts}}}} &
  \begin{minipage}[c][6em][c]{\linewidth}
    \centering
    \includegraphics[height=6em, width=0.5\linewidth]{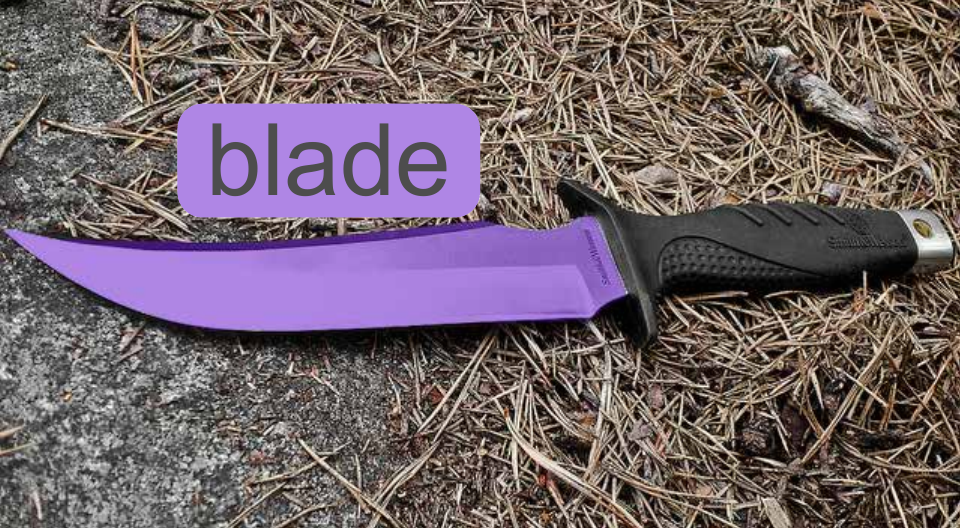}%
    \includegraphics[height=6em, width=0.5\linewidth]{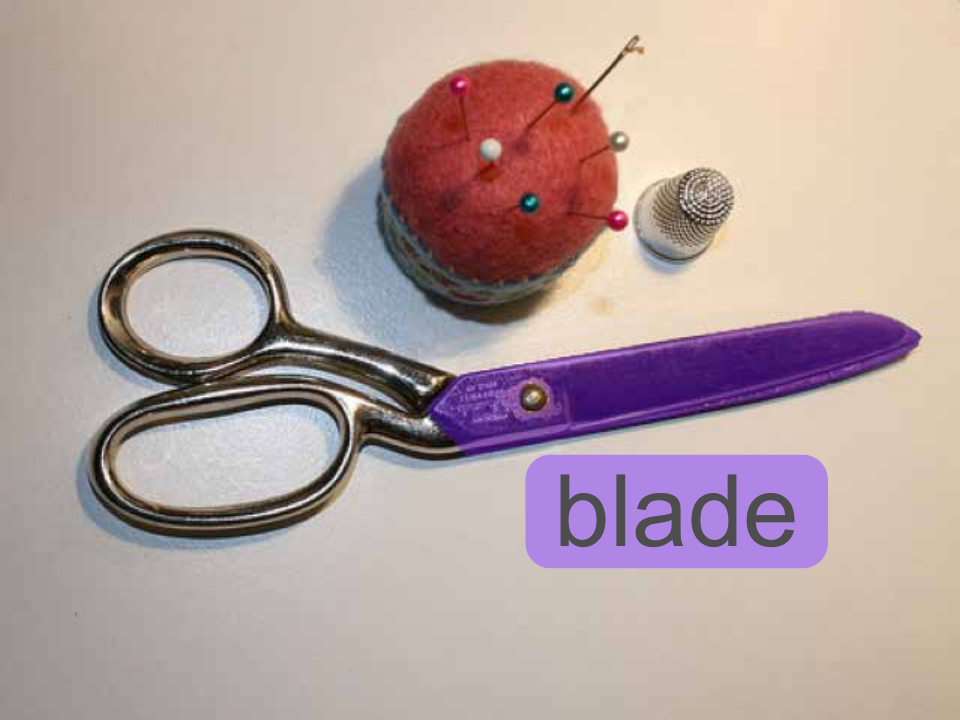}
  \end{minipage} &
  \begin{minipage}[c][6em][c]{\linewidth}
    \centering
    \includegraphics[height=6em, width=0.5\linewidth]{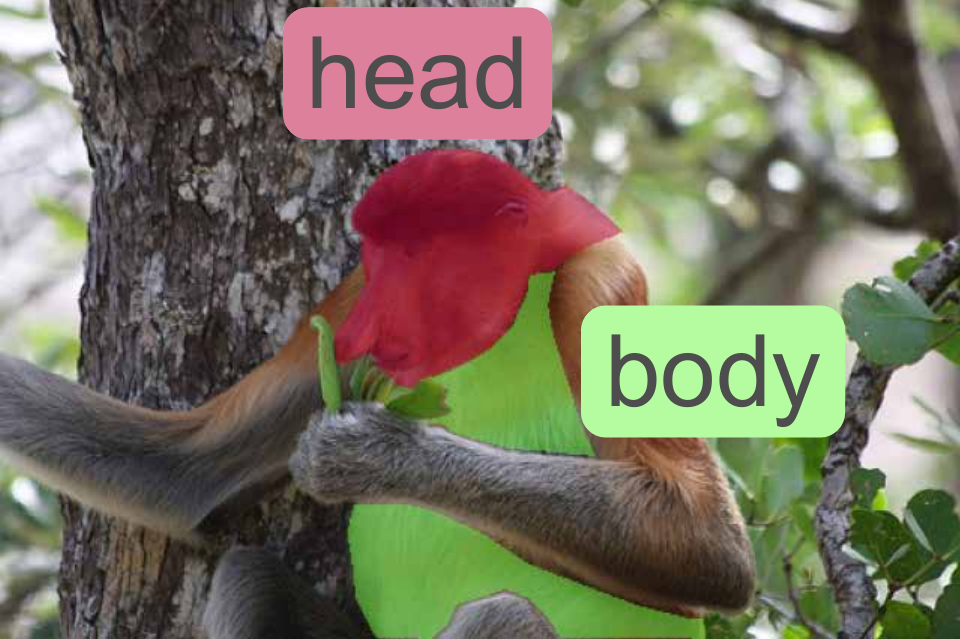}%
    \includegraphics[height=6em, width=0.5\linewidth]{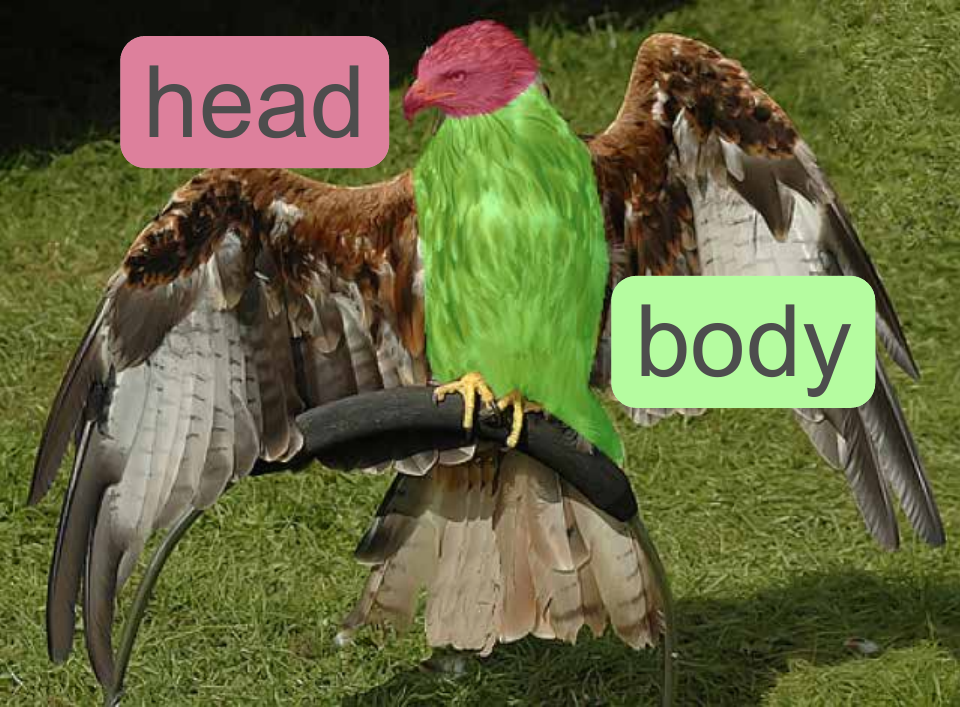}
  \end{minipage} &
  \begin{minipage}[c][6em][c]{\linewidth}
    \centering
    \includegraphics[height=6em, width=0.5\linewidth]{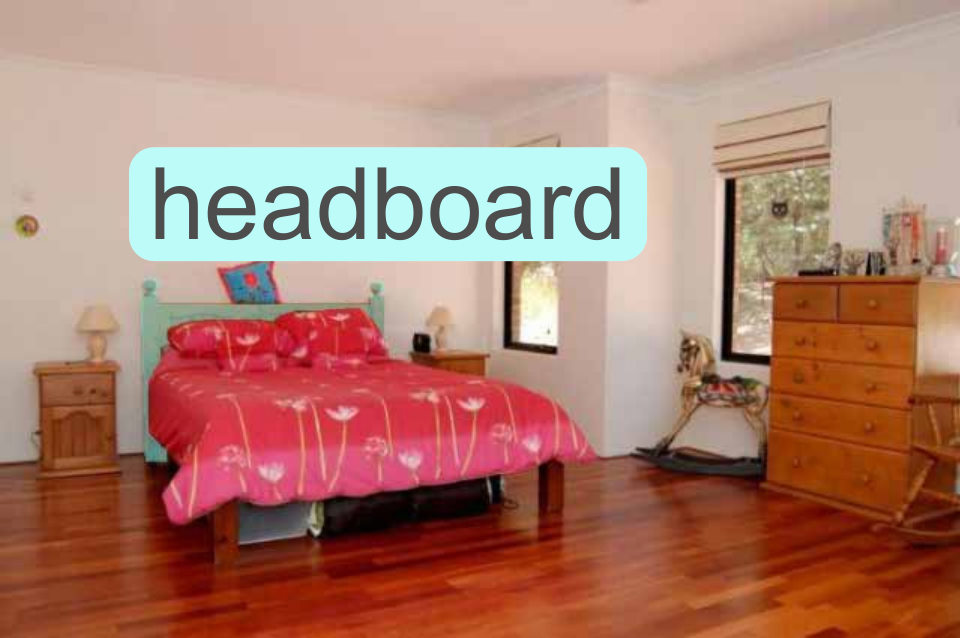}%
    \includegraphics[height=6em, width=0.5\linewidth]{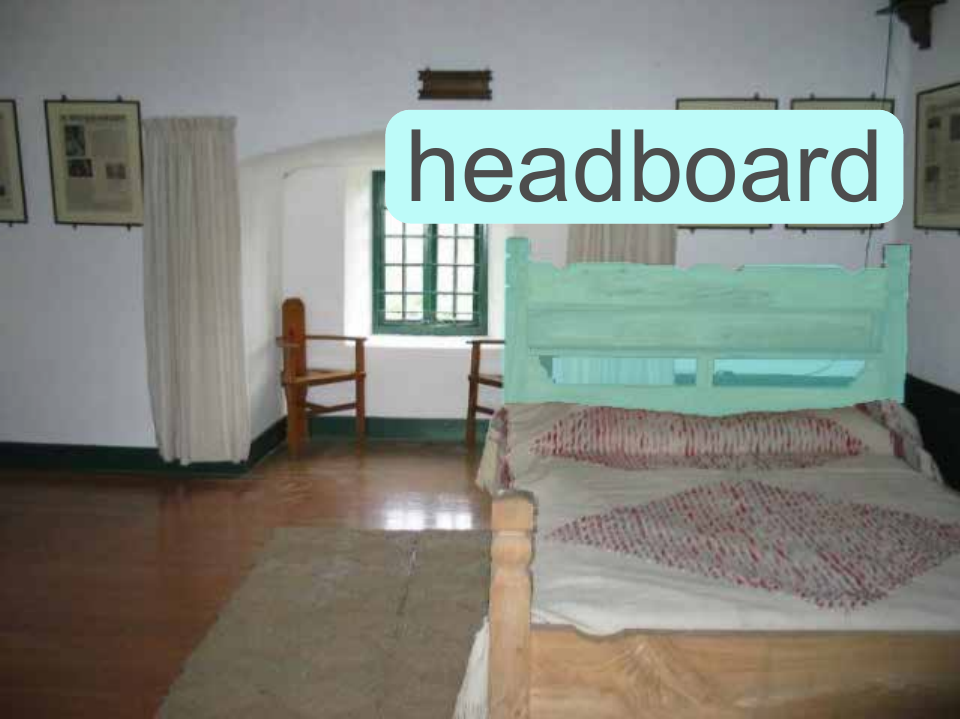}
  \end{minipage} \\
  \multirow{3}{*}{\rotatebox{90}{\parbox{0.1\textwidth}{\centering\textbf{Unique Parts}}}} &
  \begin{minipage}[c][6em][c]{\linewidth}
    \centering
    \includegraphics[height=6em, width=0.5\linewidth]{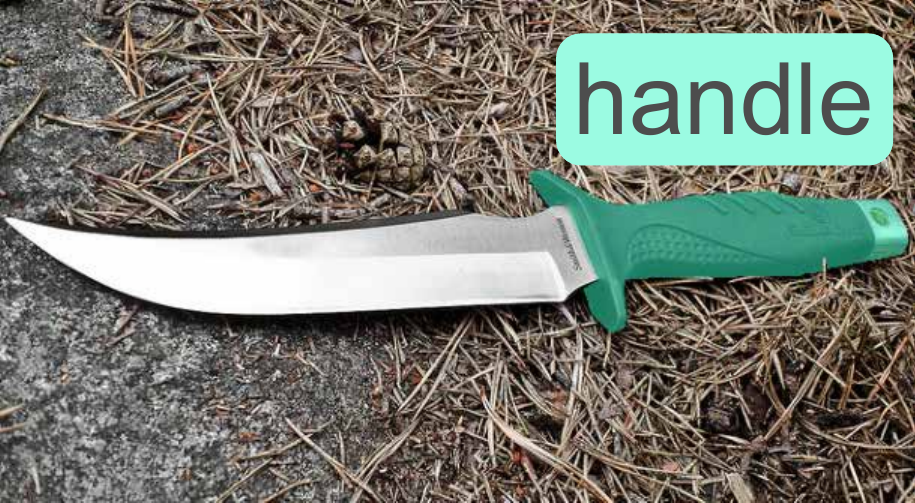}%
    \includegraphics[height=6em, width=0.5\linewidth]{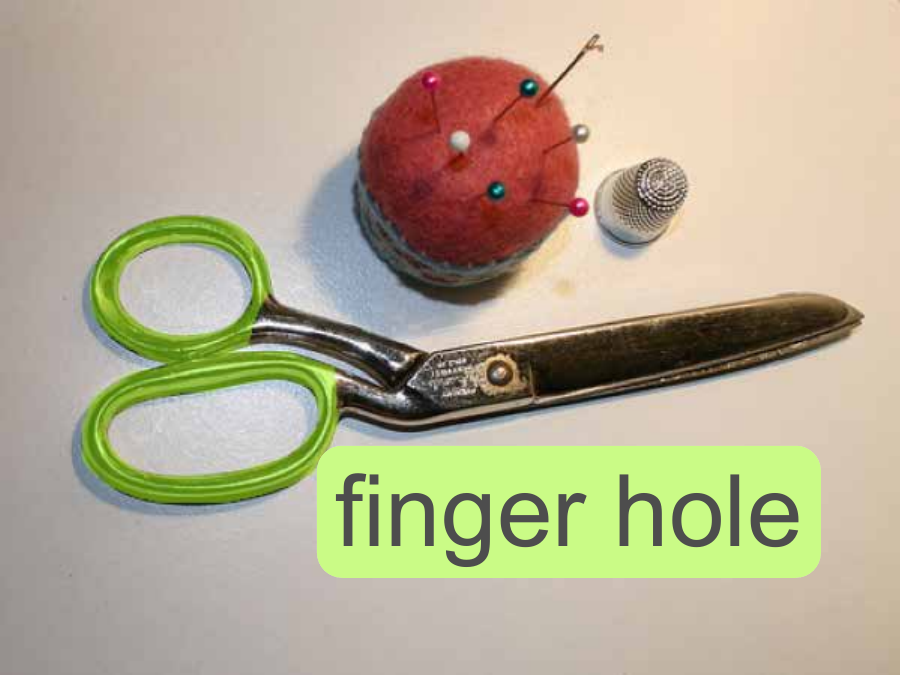}
  \end{minipage} &
  \begin{minipage}[c][6em][c]{\linewidth}
    \centering
    \includegraphics[height=6em, width=0.5\linewidth]{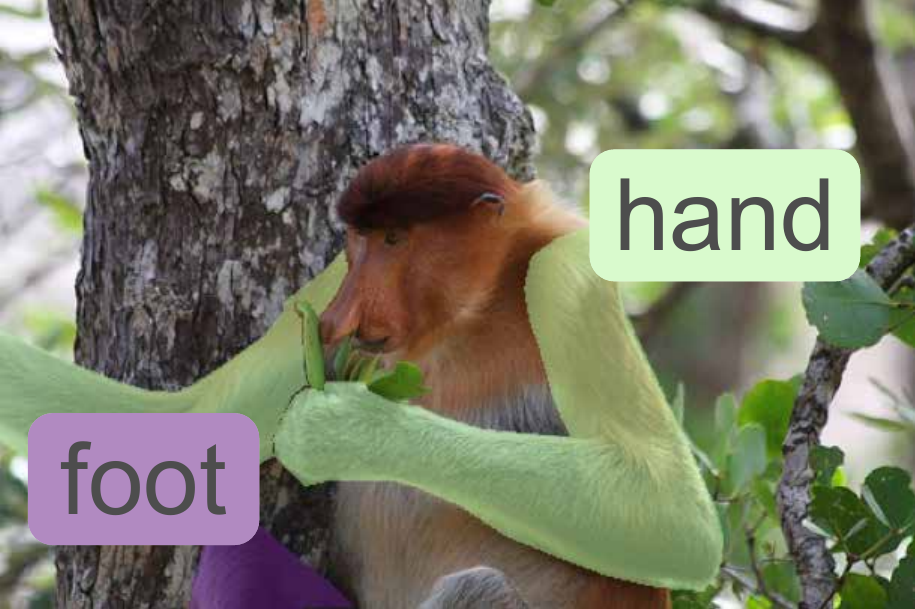}%
    \includegraphics[height=6em, width=0.5\linewidth]{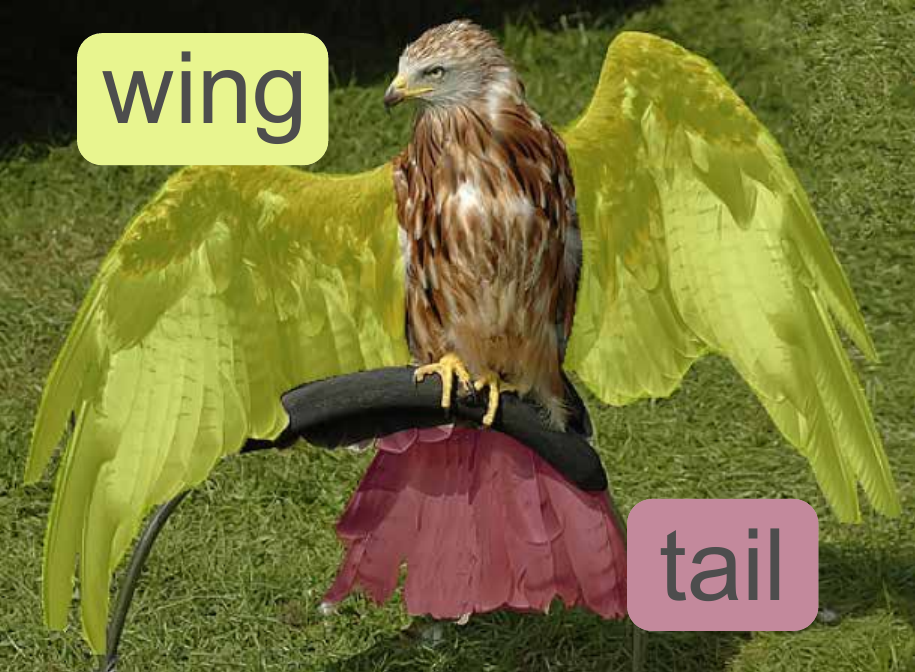}
  \end{minipage} &
  \begin{minipage}[c][6em][c]{\linewidth}
    \centering
    \includegraphics[height=6em, width=0.5\linewidth]{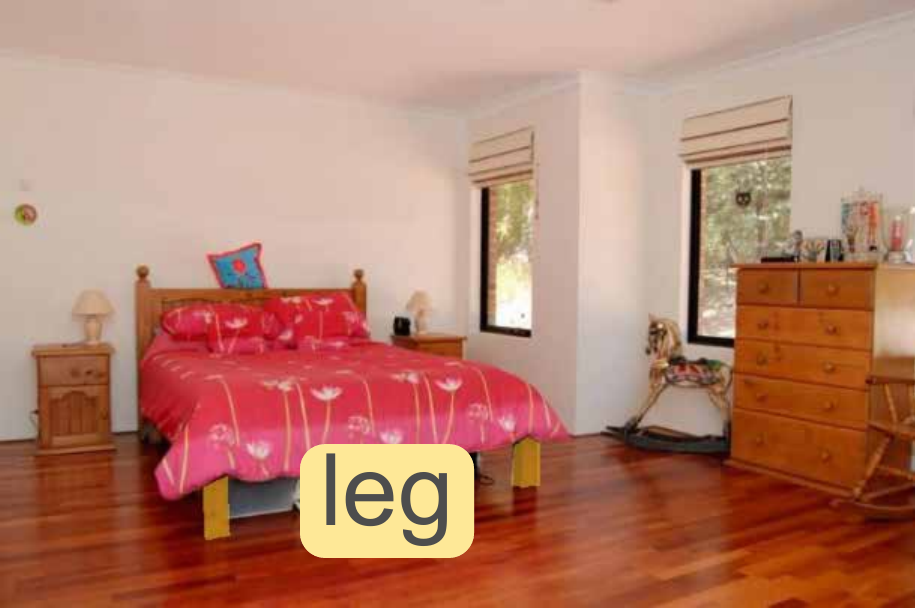}%
    \includegraphics[height=6em, width=0.5\linewidth]{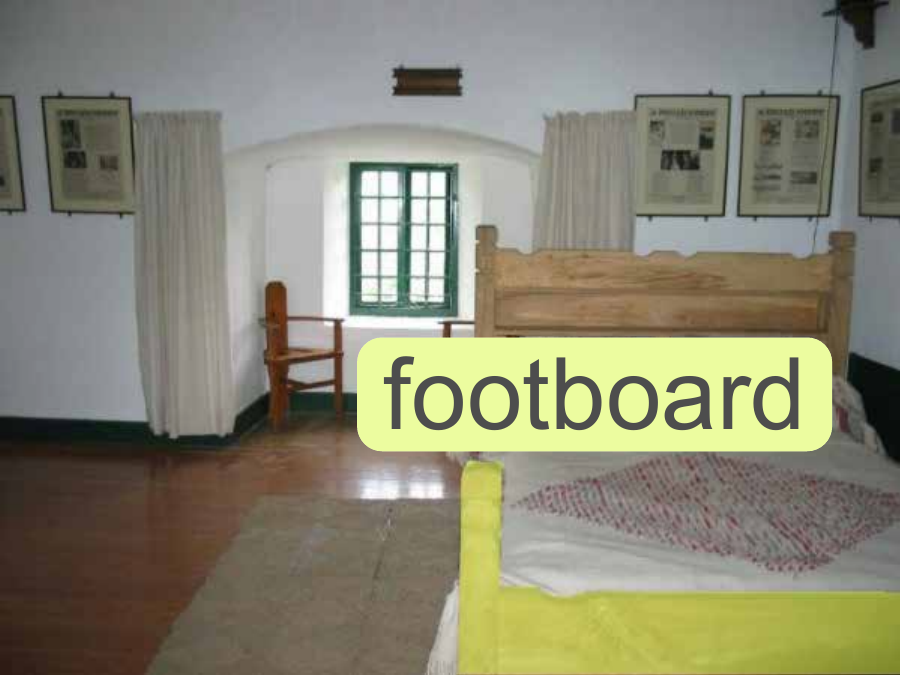}
  \end{minipage} \\
\end{tabular}
}
\vspace{-0.3cm}
\captionof{figure}{\textbf{Example Image Pairs in \benchmarkname{} with Common Objects, Common Parts, and Unique Parts} segmented and labeled. Each column represents a different image pair, derived from a set of diverse datasets with various levels of detail, PACO-LVIS, PartImageNet, and ADE20K-Part-234, covering both rigid and non-rigid objects and parts. Each image pair is displayed across 3 rows to illustrate (i) the (possibly common or different) object(s), (ii) the common object part(s), and (iii) the unique object part(s) in each pair.
}
\vspace{-0.3cm}
\label{tab:data-examples}
\end{table*}
\subsection{Correspondence Adaptation Module (CAM)}
\label{sec:correspondence-adaptation}

Due to the high cost of training LLMs with billions of parameters, many works have taken advantage of adaptive modules, with sizes merely a fraction of the LLMs' original sizes, by freezing the LLMs and only training the modules for downstream tasks~\cite{hu2021lora,zhang2023composing,huang2023lorahub,li2023fine}. These works have demonstrated strong performance and high efficiency while also avoiding catastrophic forgetting \cite{ren2024analyzing,wang2023orthogonal}. In particular, VPG-C \cite{li2023fine} effectively exhibits the capability to adapt an LLM to multi-image reasoning while only accounting for $0.09\%$ of the entire model's parameter count. Inspired by this finding, we leverage VPG-C for multi-image correspondence adaptation. Specifically, at select layers $l \in L$, we linearly project the last input token $\mathbf{t}^l_{S_T} \in \R^D$ via $f_{\textnormal{adaptation}}\!:\!\R^{D} \to \R^{D_I}$ to get an instruction-specific guidance embedding. We then use this embedding to enrich the query tokens $\mathbf{q}$ and obtain new query tokens by:
\begin{equation}
    \mathbf{q}' = \mathbf{q} + f_\text{adaptation} \left(\mathbf{t}^l_{S_T}\right) \in \R^{S_I \times D_I}.
\end{equation}
These context-guided query tokens are finally used to extract semantic- and context-rich visual information from $\mathbf{X}'_\text{global}$ from the Correspondence Extraction Module via:
\begin{equation}
    \mathbf{X}'_\text{embed} = \mathcal{Q}\left(\mathbf{q}', \mathbf{X}'_\text{global}\right) \in \R^{N_I \times S_I \times D_I}.
\end{equation}

\noindent Finally, these embeddings are projected into the language model space via $f_\text{integration}\!:\!\R^{D_I} \mapsto \R^D$ and added into the visual portions of the input $\mathbf{T}^l$ to layer $l$ of the LLM by:
\begin{equation}
    \mathbf{I}^l_\text{fused} = \mathbf{I}^l + f_\text{reintegration}\left(\mathbf{X}'_\text{embed}\right) \in \R^{N_I \times S_I \times D}.
\end{equation}

\noindent This process facilitates the integration of embeddings imbued with strong context and semantic information directly into \modelname{}. In practice, we inject our Correspondence Adaptation Modules to layers $l \in L \!=\!\{11, 22\}$, which is a third and two-thirds of the way, respectively, through a 32-layer LLM, to learn semantic correspondence at multiple granularities (\eg, object and part). Ablations on different layers $L$ in Section \ref{sec:ablation-studies} validate our design choice.

\subsection{Training Objective}
Following prior work \cite{rasheed2023glamm, lai2024lisa}, we optimize a combined training loss $\mathcal{L}\!=\!\lambda_{\textnormal{text}}\mathcal{L}_{\textnormal{text}} + \mathcal{L}_{\textnormal{mask}}$, 
where $\mathcal{L}_{\text{text}}$ is the next-token prediction loss and $\mathcal{L}_{\text{mask}}$ is the segmentation loss. The weights $\lambda_{\text{text}}$, $\lambda_{\text{focal}}$, and $\lambda_{\text{Dice}}$ control the contribution of each loss component.
Here, $\mathcal{L}_{\textnormal{text}}$ is a causal cross-entropy (CE) loss computed from the predicted and right-shifted ground truth tokens $\mathbf{\hat{T}}^K$ and $\mathbf{y}$, 
\begin{equation}
\scalebox{0.9}{$
    \mathcal{L}_{\textnormal{text}} = \textnormal{CE}\left(\mathbf{\hat{T}}^K, \mathbf{y}\right),
    $}
\end{equation}
while $\mathcal{L}_{\textnormal{mask}}$ combines a focal loss \cite{ross2017focal} and a DICE loss \cite{milletari2016v} derived from predicted and ground truth masks $\mathbf{\hat{M}}$ and $\mathbf{M}$, 
\begin{equation}
\scalebox{0.9}{$
    \mathcal{L}_{\textnormal{mask}} = \lambda_{\textnormal{focal}}\mathcal{L}_\textnormal{focal}\left(\mathbf{\hat{M}}, \mathbf{M}\right) + \lambda_{\textnormal{Dice}} \mathcal{L}_\textnormal{Dice}\left(\mathbf{\hat{M}}, \mathbf{M}\right).
$}
\end{equation}
\section{\benchmarkname{} Dataset}
Although multi-image datasets of various scales are available, they present combinations of limitations that make them unsuitable for the part-focused semantic co-segmentation task. These include the absence of fine-grained masks for segmentation \cite{huang2023sparkles, li2023fine, suhr2017corpus, suhr2018corpus, lu2021iconqa}, datasets being too small or domain-specific to facilitate generalizable LVLM training despite containing localized labels \cite{batra2010icoseg, faktor2013co, shotton2006textonboost, rubinstein2013unsupervised, welinder2010caltech}, or the lack of part-level information altogether \cite{pramanick2024jack, rubinstein2013unsupervised}. 
To address these gaps, we introduce \benchmarkname{}, a new dataset curated from publicly available sources to support training and evaluation of part-focused semantic co-segmentation models. Figure \ref{tab:data-examples} provides examples from our dataset, highlighting its diversity in object categories and visual details. 
Appendix~\ref{appendix:mixedparts} describes the dataset construction process and source datasets, while Appendix~\ref{appendix:datasets} summarizes dataset statistics.
\section{Experiments}

\begin{table*}[t!]
    \centering
    \resizebox{\linewidth}{!}{
        \begin{tabulary}{\textwidth}{>{\centering\arraybackslash}p{.29\textwidth} >{\centering\arraybackslash}p{.28\textwidth} >{\centering\arraybackslash}p{.29\textwidth} >{\centering\arraybackslash}p{.28\textwidth} >{\centering\arraybackslash}p{.3\textwidth} >{\centering\arraybackslash}p{.3\textwidth}}
        \toprule
            \multicolumn{2}{>{\centering\arraybackslash}p{.57\textwidth}}{(a) \textbf{Common Object}} & \multicolumn{2}{>{\centering\arraybackslash}p{.57\textwidth}}{(b) \textbf{Common Parts}} & \multicolumn{2}{>{\centering\arraybackslash}p{.6\textwidth}}{(c) \textbf{Unique Parts}} \\
            \midrule
            \noalign{\vskip 2pt} 
            \includegraphics[width=\linewidth, height=4cm]{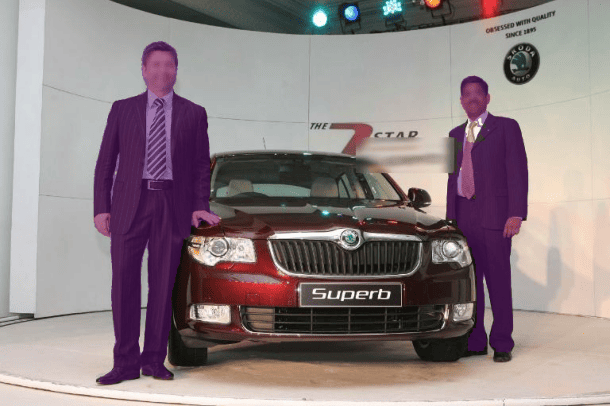} & 
            \includegraphics[width=\linewidth, height=4cm]{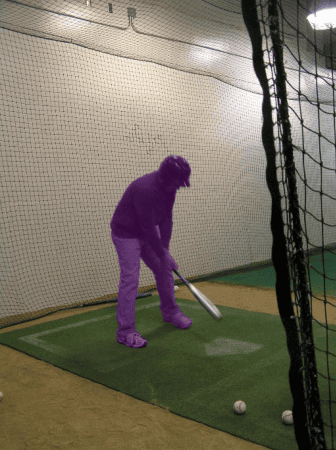} & 
            \includegraphics[width=\linewidth, height=4cm]{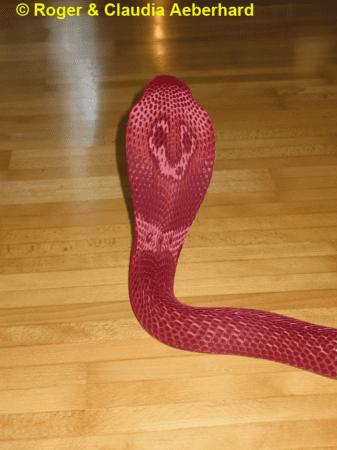} & 
            \includegraphics[width=\linewidth, height=4cm]{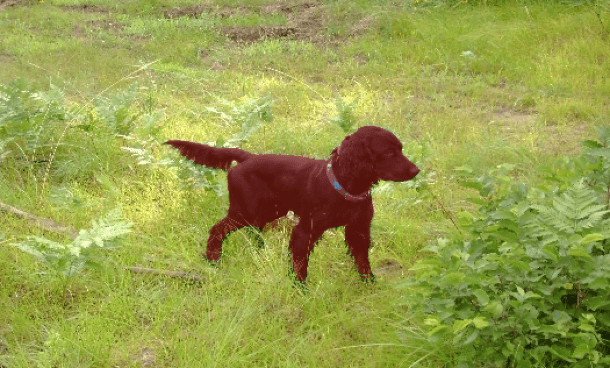} &
            \includegraphics[width=\linewidth, height=4cm]{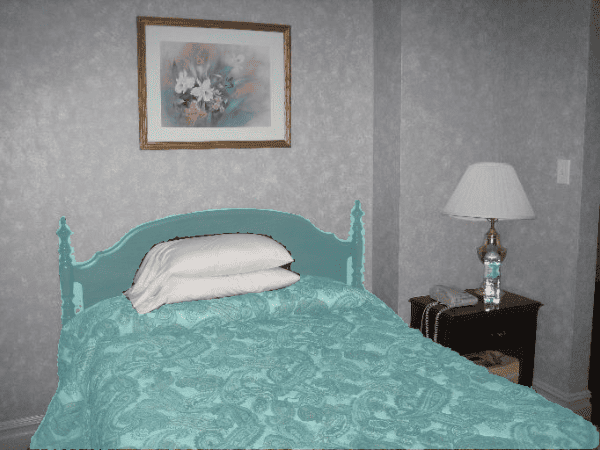} & 
            \includegraphics[width=\linewidth, height=4cm]{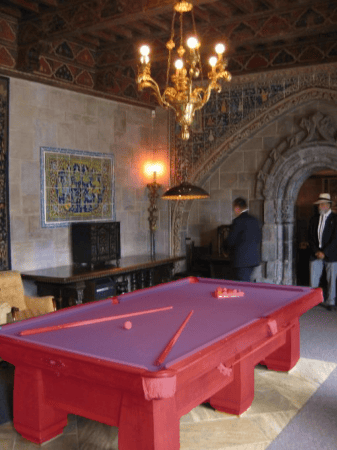}  \\
            \multicolumn{2}{p{.57\textwidth}}{The common object is \colorbox{object1!50}{the person}.} 
            & \multicolumn{2}{p{.57\textwidth}}{The images show \colorbox{common1!50}{a snake} and \colorbox{common2!50}{a dog}.} & \multicolumn{2}{p{.57\textwidth}}{The images show \colorbox{unique1!80}{a bed} and \colorbox{unique2!50}{a pool table}.} \\
            \cline{1-2}
            \noalign{\vskip 3pt}  
            \includegraphics[width=\linewidth, height=4cm]{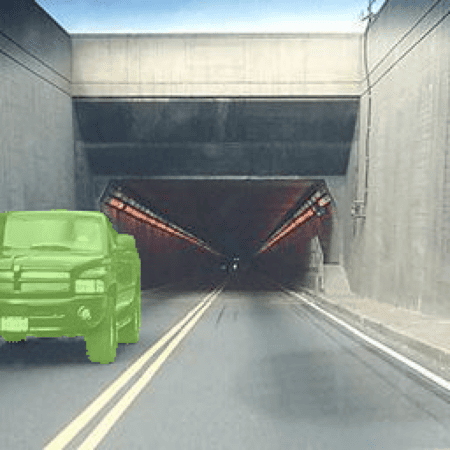} & 
            \includegraphics[width=\linewidth, height=4cm]{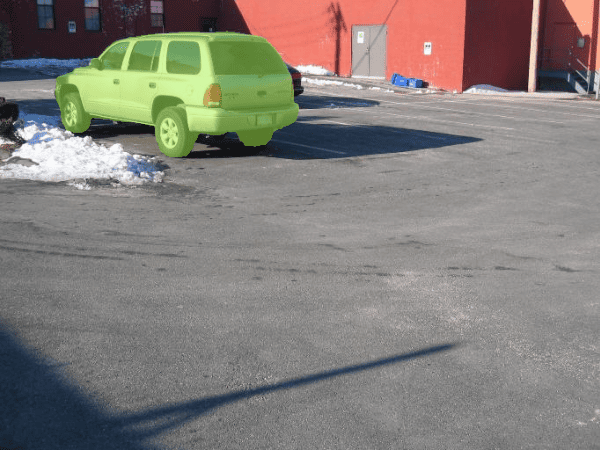} & 
            \includegraphics[width=\linewidth, height=4cm]{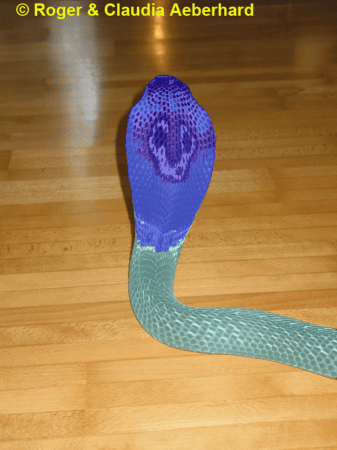} & 
            \includegraphics[width=\linewidth, height=4cm]{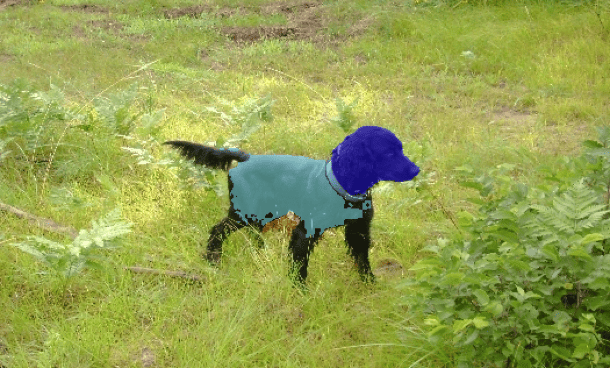} &
            \includegraphics[width=\linewidth, height=4cm]{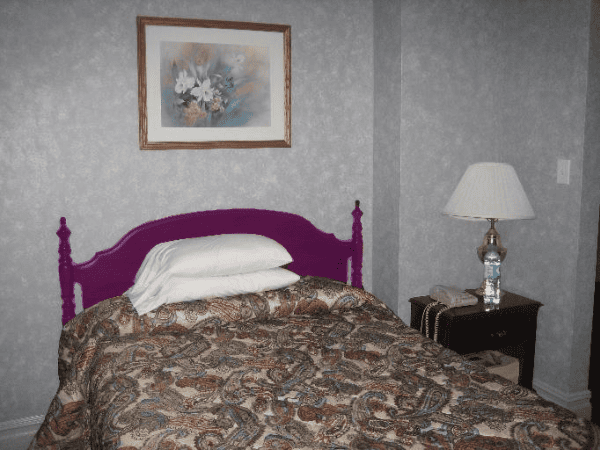} & 
            \includegraphics[width=\linewidth, height=4cm]{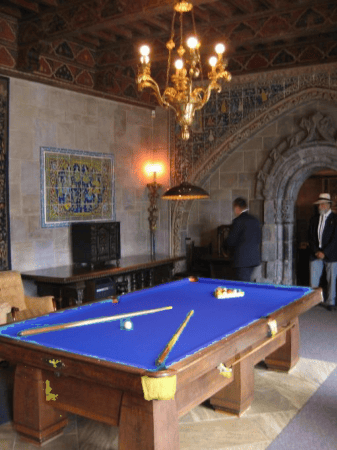}  \\
            \multicolumn{2}{p{.57\textwidth}}{The images include \colorbox{object2!50}{a car}.} 
            & \multicolumn{2}{p{.57\textwidth}}{The detected common parts are \colorbox{common3!80}{a body} and \colorbox{common4!40}{a head}.} & \multicolumn{2}{p{.6\textwidth}}{The unique parts present are \colorbox{unique3!50}{a headboard}, \colorbox{unique4!50}{a bed}, and \colorbox{unique5!80}{a pocket}.} \\
        \bottomrule
        \end{tabulary}
    }
    \vspace{-0.3cm}
    \captionof{figure}{\textbf{Qualitative Results.} \modelname{} can co-segment (a) common objects, (b) shared parts between objects from different classes, and (c) distinct parts unique to each object. Our model demonstrates pixel-grounded understanding of various non-rigid (\eg, humans, animals) and rigid objects (\eg, car, bed), including less common objects and their parts (\eg, the bed and pocket of a pool table).}
    \label{tab:qualitative_results}
    \vspace{-0.3cm}
\end{table*}
\noindent \textbf{Metrics.}
We evaluate \modelname{}'s performance on the \benchmarkname{} dataset, reporting the Average Precision (\textbf{AP50}), mean Intersection over Union (\textbf{mIoU}), and \textbf{Recall}, which assess the model's segmentation performance. To evaluate its semantic label generation capability, following existing works \cite{yuan2023osprey,conti2024vocabulary}, we employ Semantic Similarity (\textbf{SS}) and Semantic IoU (\textbf{S-IoU}). Implementation details and descriptions of the evaluation metrics can be found in Appendix \ref{appendix:eval-metrics}. We perform evaluation on $\sim$1K image pairs, ensuring an equal distribution of image pairs from each original dataset and maintaining equal representation across all tasks. We report performance on all three subtasks of part-focused semantic co-segmentation -- common objects, common parts, unique parts -- and their average, which reflects overall performance on the \benchmarkname{} test set.

\noindent \textbf{Baselines.}
To the best of our knowledge, this is the first effort to tackle multi-image part-focused co-segmentation with part label identification. There is thus a lack of baselines for this new task, which we rectify by designing our own baselines from strong pretrained models in the semantic segmentation literature.
Our baselines comprise three zero-shot approaches: one modular method (\textbf{Cascade}) that combines pretrained models \cite{huang2023sparkles, achiam2023gpt, lai2024lisa}, each excelling at a different aspect of our novel task, and two methods that leverage state-of-the-art part segmentation models to compare their outputs (\textbf{PartGLEE} \cite{li2024partglee} and \textbf{VLPart} \cite{sun2023going}). Additionally, we report results for two finetuned segmentation-based LVLMs (\textbf{GLaMM} \cite{rasheed2023glamm} and \textbf{LISA} \cite{lai2024lisa}). Implementation details for all models are available in Appendix \ref{appendix:impl}.

\begin{table}[t!]
\centering
\resizebox{\columnwidth}{!}{
\begin{tabular}{l|ccc|cc}
\toprule
\textbf{Method} & AP50 & mIoU & Recall & SS & S-IoU \\ \hline \hline
Cascade \cite{huang2023sparkles, achiam2023gpt, lai2024lisa} & 5.7 & 27.9 & 19.0 & 32.2 & 14.8 \\
Multi-Image PartGLEE \cite{li2024partglee} & 1.2 & 29.3 & 9.7 & 78.5 & 63.3 \\
Multi-Image VLPart \cite{sun2023going} & 13.4 & 42.8 & 34.6 & 59.1 & 46.5 \\
\hline
Multi-Image GLaMM \cite{rasheed2023glamm} & 42.9 & 59.9 & 54.9 & 76.8 & 71.2 \\
Multi-Image LISA \cite{lai2024lisa} & 41.4 & 59.7 & 55.5 & 78.7 & 72.5 \\
\hline
\cc \textbf{\modelname{}} & \cc \textbf{45.9} & \cc \textbf{63.7} & \cc \textbf{59.7} & \cc \textbf{82.7} & \cc \textbf{77.1} \\
\bottomrule
\end{tabular}
}
\vspace{-0.3cm}
\caption{\textbf{Experimental Results on \benchmarkname{}}. The first three metrics are segmentation-based, while the last two are text-based. CALICO outperforms baselines across all metrics.}
\label{tab:mixedparts-results}
\vspace{-0.4cm}
\end{table}

\subsection{Experimental Results}

Table \ref{tab:mixedparts-results} presents results comparing \modelname{} against our baselines. \modelname{} demonstrates superior performance on all metrics compared to all zero-shot and finetuned approaches, achieving relative gains of 7.0\%, 6.3\%, and 7.6\% on segmentation-based metrics AP50, mIoU, and Recall, respectively, and 5.0\% and 6.3\% on text-based metrics SS and S-IoU. These performance gains come with no overhead; in fact, as shown in Figure \ref{fig:tflops}, \modelname{} demonstrates high efficiency compared to LISA and GLaMM in terms of TFLOPS and inference time. Details can be found in Appendix \ref{appendix:efficiency}. 

The cascade method fails to perform well on our task, highlighting the task’s challenging nature and suggesting that it cannot be addressed merely by combining standalone modules. Although PartGLEE and VLPart demonstrate strong zero-shot single-image object-part segmentation performance, they struggle with our multi-image tasks. In particular, PartGLEE produces 29 object predictions per image on average, increasing its likelihood of matching ground truth labels; yet \modelname{}'s text scores are superior despite focusing on a single most relevant object in context. When finetuned on \benchmarkname{}, Multi-Image GLaMM and LISA exhibit performance improvements compared to the zero-shot baselines. However, GLaMM still lags behind \modelname{}'s performance despite both initialized from the same weights, demonstrating the effectiveness of our proposed modules, which we further ablate in Section \ref{sec:ablation-studies}.

Qualitative results in Figure \ref{tab:qualitative_results} showcase \modelname{}'s strong pixel-level understanding across all three tasks in part-focused semantic co-segmentation. Our model effectively localizes both non-rigid objects (\eg, humans, animals) and rigid ones (\eg, cars, beds), and demonstrates part-level reasoning even for less common object classes and their parts, such as the bed and pocket of a pool table. Furthermore, Figure \ref{tab:calico_in_context} illustrates \modelname{}'s contextual understanding, where different image pairings prompt the model to segment different objects accordingly, rather than defaulting to the most salient object in each image. Additional experiments are detailed in Appendix \ref{appendix:additional-experiments}.

\begin{table}[t!]
    \centering
     \resizebox{0.93\columnwidth}{!}{
        \begin{tabulary}{\linewidth}{>{\centering\arraybackslash}p{.4\linewidth} >{\centering\arraybackslash}p{.58\linewidth}} \toprule
            \includegraphics[width=\linewidth, height=3cm]{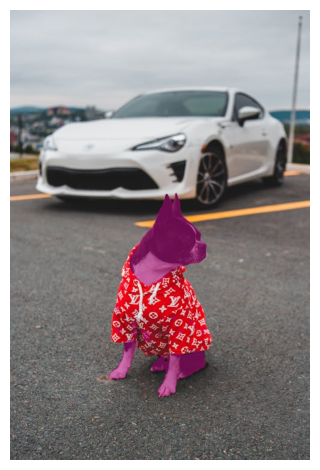} & 
            \includegraphics[width=\linewidth, height=3cm]{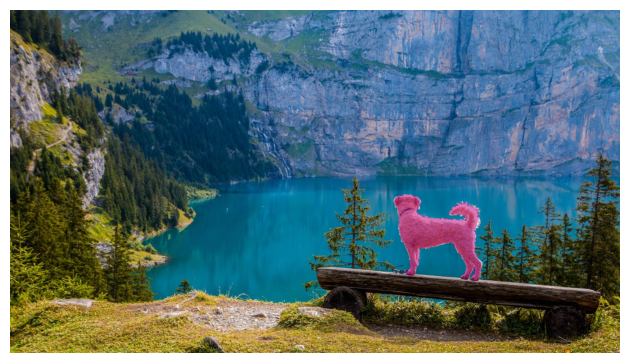} \\
            \multicolumn{2}{p{\linewidth}}{The images show \colorbox{incontext1!50}{a dog}.} \\
            \midrule
            \includegraphics[width=\linewidth, height=3cm]{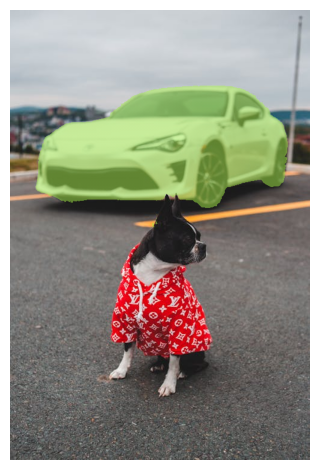} & 
            \includegraphics[width=\linewidth, height=3cm]{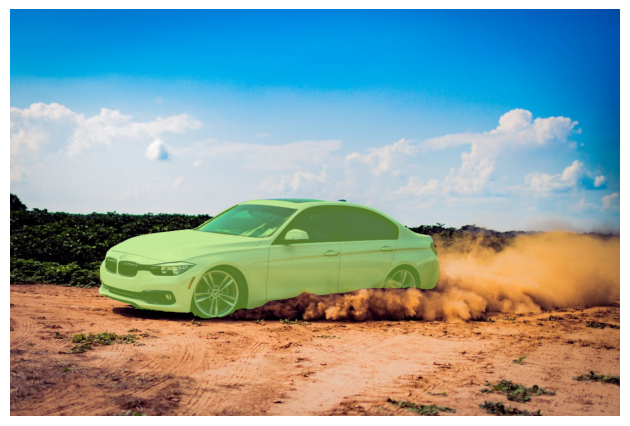} \\
            \multicolumn{2}{p{\linewidth}}{The images include \colorbox{incontext2!50}{a car}.} \\
            \bottomrule
        \end{tabulary}
    }
    \vspace{-0.3cm}
    \captionof{figure}{\textbf{CALICO In Context.} \modelname{} outputs are highly context-driven when distinguishing objects across images, despite variations in angle, size, saliency, \etc}
    \label{tab:calico_in_context}
    \vspace{-0.5cm}
\end{table}

\subsection{Ablations}
\label{sec:ablation-studies}
\noindent \textbf{\modelname{} Components.} 
We perform ablations to assess the contribution of each component in \modelname{}, including the Correspondence Extraction Module (CEM) and the Correspondence Adaptation Module (CAM), alongside Q-Former and DINOv2. Table~\ref{tab:ablations_modules} reports results for \modelname{}  variants: without Q-Former (w/o Q-Former), without DINOv2 in CEM (w/o DINO), without CEM (w/o CEM), without CAMs (w/o CAM), or removing both (w/o CEM w/o CAM). Specifically, the w/o CEM variant fuses each LVLM layer's image embeddings with only the previous layer’s last hidden state, while w/o CAM injects CEM features without fusing them with the last hidden state. In the w/o DINO setting, no cross-attention is applied, as there are no DINOv2 features to fuse with EVA-CLIP.
Since our CEM and CAM modules are designed for models with Q-Formers, we approximate the w/o Q-Former variant using a linear projection, similar to the linear ablation in \cite{li2023fine}. 

Ablation results show that each proposed module
contributes to \modelname{}'s ability to accurately co-segment and identify common objects, common parts, and unique parts. Firstly, removing Q-Former -- and thus disabling most of CAM and CEM -- leads to a substantial drop in performance compared to \modelname{}, underscoring its importance within our model architecture. Removing CEM results in a decrease in performance across all metrics relative to \modelname{} and demonstrates comparable performance to the ``w/o DINO'' variant, highlighting the critical role of both DINOv2 and CEM in \modelname{}.
CEM substantially contributes to segmentation performance as this module enables the model to learn semantic relationships in multi-image contexts.
Excluding the CAM modules leads to a decline in labeling performance, suggesting that CAM facilitates the integration of image-derived features into the language space. Interestingly, keeping CAM while removing CEM results in worse performance than removing both, indicating that CAM may introduce redundant or uninformative features when not guided by external signals from CEM. In contrast, when CEM is present, removing CAM still outperforms the variant without both modules, and full \modelname{} achieves the best results. This demonstrates that CAM is most beneficial when there are external semantic signals for the part-focused semantic co-segmentation task.

\noindent \textbf{\modelname{} Injecting CAMs.} We perform further ablations to examine the efficacy of injecting 2 evenly spaced Correspondence Adaptation Modules (CAM) into our \modelname{} LVLM, and present results in Figure \ref{fig:ablation-layers}. Past works \cite{xin2020deebert,li2023fine} have demonstrated the effectiveness of using or injecting information at the intermediate layer $\frac{K}{2}$ as guidance for learning. Since our task involves multimodal understanding at multiple granularities (object and part), we use 2 evenly spaced layers to incorporate semantic features at different levels of the LVLM's learning, encouraging the model to focus on various object-part correspondences. This results in the best segmentation and labeling performance compared to the 1-layer or 3-layer variants.
\begin{figure}[t!]
\centering
    \resizebox{0.95\columnwidth}{!}{
        \begin{tikzpicture}
            \begin{axis}[
                ybar=0pt,
                ymin=0,
                width=1.2\columnwidth,
                height=5cm,
                enlarge x limits=0.3,
                ymajorgrids, tick align=inside,
                symbolic x coords={16, 11+22 (CALICO), 8+16+24},
                xtick=data,
                nodes near coords,
                every node near coord/.append style={
                    font=\footnotesize,
                    text width=1cm,
                    rotate=90,
                    align=center,
                    xshift=11pt,
                    yshift=-6pt,
                    color=black,
                    /pgf/number format/.cd,
                        fixed,
                        fixed zerofill,
                        precision=1
                },
                ylabel={Metric Scores (\%)},
                axis x line*=bottom,
                axis y line*=left,
                ylabel style={align=center, yshift=-4pt},
                y tick label style={
                    /pgf/number format/assume math mode},
                bar width=0.35cm,
                legend style={
                                at={(0.5,1.35)},
                                anchor=north,
                                legend columns=-1,
                                /tikz/every even column/.append style={column sep=0.5cm}
                            },
                legend image code/.code={%
                      \draw[#1] (0cm,-0.1cm) rectangle (0.6cm,0.1cm);
                    },
            ]
            \addplot[blue,fill=blue!40] coordinates {(16,44.7) (11+22 (CALICO),45.9) (8+16+24,43.8) };
            \addplot[red,fill=red!40] coordinates {(16,63.2) (11+22 (CALICO),63.7) (8+16+24,61.4)};
            \addplot[purple,fill=purple!40] coordinates {(16,58.6) (11+22 (CALICO),59.7) (8+16+24,57.2)};
            \addplot[orange,fill=orange!40] coordinates {(16,82.3) (11+22 (CALICO),82.7) (8+16+24,79.7)};
            \addplot[green,fill=green!40] coordinates {(16,76.8) (11+22 (CALICO),77.1) (8+16+24,74.3)};
            \legend{AP50, mIoU, Recall, SS, S-IoU}
            \end{axis}
        \end{tikzpicture}
    }
\vspace{-0.3cm}
\caption{\textbf{Ablations on CAM layer injection.}}
\label{fig:ablation-layers}
\end{figure}
\begin{table}[t!]
\centering
\resizebox{0.99\columnwidth}{!}{
\begin{tabular}{l|ccc|cc}
\toprule
    \textbf{Setting} & AP50 & mIoU & Recall & SS & S-IoU \\ \hline \hline
    w/o Q-Former & 38.5 & 59.2 & 44.8 & 64.5 & 55.6 \\
    w/o DINO & 43.9 & 61.7 & 57.1 & 80.2 & 74.5 \\
    w/o CEM & 43.6 & 61.6 & 57.5 & 80.8 & 75.2 \\
    w/o CAM & \textbf{45.9} & 63.3 & \textbf{59.7} & 82.0 & 76.5 \\
    w/o CEM w/o CAM & 44.1 & 62.7 & 58.1 & 81.6 & 76.3 \\
    \hline
    \cc \textbf{\modelname{}} & \cc \textbf{45.9} & \cc \textbf{63.7} & \cc \textbf{59.7} & \cc \textbf{82.7} & \cc \textbf{77.1} \\
    \bottomrule
\end{tabular}
}
\vspace{-0.3cm}
\caption{\textbf{Ablations on \modelname{} Components}, including Q-Former, DINOv2, the Correspondence Extraction Module, and the Correspondence Adaptation Modules.}
\label{tab:ablations_modules}
\vspace{-0.3cm}
\end{table}

\section{Conclusion}
This paper introduces the novel task of part-focused semantic co-segmentation, which involves the segmentation of common and unique parts across multiple images, laying the groundwork for future research in
enhancing the capability of Large Vision-Language Models (LVLMs) to analyze and interpret complex visual data in a granular manner. To solve this task, we propose \modelname{}, an LVLM that incorporates a novel correspondence extraction module and an adaptation module to handle multi-image part relationships. Experiments conducted on the newly curated \benchmarkname{} dataset demonstrate that \modelname{} can effectively identify and segment common/unique parts with high accuracy, outperforming existing models.

\section*{Acknowledgments}
This research is based upon work supported by U.S. DARPA ECOLE Program No. HR00112390062 and U.S. DARPA SciFy Program No. HR001125C0303. The views and conclusions contained herein are those of the authors and should not be interpreted as necessarily representing the official policies, either expressed or implied, of DARPA or the U.S. Government. The U.S. Government is authorized to reproduce and distribute reprints for governmental purposes notwithstanding any copyright annotation therein. This work used the Delta GPUs provided by the National Center for Supercomputing Applications through allocations CIS240752, CIS240817, and CIS250130 from the Advanced Cyberinfrastructure Coordination Ecosystem: Services \& Support (ACCESS, \citet{boerner2023access}) program, supported by National Science Foundation grants \#2138259, \#2138286, \#2138307, \#2137603, and \#2138296.

{
    \small
    \bibliographystyle{ieeenat_fullname}
    \bibliography{main}

\begin{thebibliography}{81}
\providecommand{\natexlab}[1]{#1}
\providecommand{\url}[1]{\texttt{#1}}
\expandafter\ifx\csname urlstyle\endcsname\relax
  \providecommand{\doi}[1]{doi: #1}\else
  \providecommand{\doi}{doi: \begingroup \urlstyle{rm}\Url}\fi

\bibitem[Achiam et~al.(2023)Achiam, Adler, Agarwal, Ahmad, Akkaya, Aleman, Almeida, Altenschmidt, Altman, Anadkat, et~al.]{achiam2023gpt}
Josh Achiam, Steven Adler, Sandhini Agarwal, Lama Ahmad, Ilge Akkaya, Florencia~Leoni Aleman, Diogo Almeida, Janko Altenschmidt, Sam Altman, Shyamal Anadkat, et~al.
\newblock {GPT-4} technical report.
\newblock \emph{arXiv preprint arXiv:2303.08774}, 2023.

\bibitem[Amir et~al.(2022)Amir, Gandelsman, Bagon, and Dekel]{amir2021deep}
Shir Amir, Yossi Gandelsman, Shai Bagon, and Tali Dekel.
\newblock Deep {ViT} features as dense visual descriptors.
\newblock In \emph{Workshop Proceedings of the European Conference on Computer Vision (ECCVW) What is Motion For?}, 2022.

\bibitem[Batra et~al.(2010)Batra, Kowdle, Parikh, Luo, and Chen]{batra2010icoseg}
Dhruv Batra, Adarsh Kowdle, Devi Parikh, Jiebo Luo, and Tsuhan Chen.
\newblock {iCoseg}: Interactive co-segmentation with intelligent scribble guidance.
\newblock In \emph{Proceedings of the IEEE Computer Society Conference on Computer Vision and Pattern Recognition}, 2010.

\bibitem[Boerner et~al.(2023)Boerner, Deems, Furlani, Knuth, and Towns]{boerner2023access}
Timothy~J Boerner, Stephen Deems, Thomas~R Furlani, Shelley~L Knuth, and John Towns.
\newblock {ACCESS}: Advancing innovation: Nsf's advanced cyberinfrastructure coordination ecosystem: Services \& support.
\newblock In \emph{2023 Practice and Experience in Advanced Research Computing, PEARC 2023}, pages 173--176. Association for Computing Machinery, 2023.

\bibitem[Caron et~al.(2021)Caron, Touvron, Misra, J{\'e}gou, Mairal, Bojanowski, and Joulin]{caron2021emerging}
Mathilde Caron, Hugo Touvron, Ishan Misra, Herv{\'e} J{\'e}gou, Julien Mairal, Piotr Bojanowski, and Armand Joulin.
\newblock Emerging properties in self-supervised vision transformers.
\newblock In \emph{Proceedings of the IEEE/CVF International Conference on Computer Vision (CVPR)}, 2021.

\bibitem[Cho et~al.(2023)Cho, Kr{\"a}henb{\"u}hl, and Ramanathan]{cho2023partdistillation}
Jang~Hyun Cho, Philipp Kr{\"a}henb{\"u}hl, and Vignesh Ramanathan.
\newblock {PartDistillation}: Learning parts from instance segmentation.
\newblock In \emph{Proceedings of the IEEE/CVF Conference on Computer Vision and Pattern Recognition (CVPR)}, 2023.

\bibitem[Choudhury et~al.(2021)Choudhury, Laina, Rupprecht, and Vedaldi]{choudhury2021unsupervised}
Subhabrata Choudhury, Iro Laina, Christian Rupprecht, and Andrea Vedaldi.
\newblock Unsupervised part discovery from contrastive reconstruction.
\newblock In \emph{Proceedings of the Conference on Neural Information Processing Systems (NeurIPS)}, 2021.

\bibitem[Collins et~al.(2018)Collins, Achanta, and Susstrunk]{collins2018deep}
Edo Collins, Radhakrishna Achanta, and Sabine Susstrunk.
\newblock Deep feature factorization for concept discovery.
\newblock In \emph{Proceedings of the European Conference on Computer Vision (ECCV)}, 2018.

\bibitem[Conti et~al.(2023)Conti, Fini, Mancini, Rota, Wang, and Ricci]{conti2024vocabulary}
Alessandro Conti, Enrico Fini, Massimiliano Mancini, Paolo Rota, Yiming Wang, and Elisa Ricci.
\newblock Vocabulary-free image classification.
\newblock In \emph{Proceedings of the Conference on Neural Information Processing Systems (NeurIPS)}, 2023.

\bibitem[Dai et~al.(2023)Dai, Li, Li, Tiong, Zhao, Wang, Li, Fung, and Hoi]{dai2023instructblip}
Wenliang Dai, Junnan Li, Dongxu Li, Anthony Meng~Huat Tiong, Junqi Zhao, Weisheng Wang, Boyang Li, Pascale Fung, and Steven Hoi.
\newblock {InstructBLIP}: Towards general-purpose vision-language models with instruction tuning, 2023.

\bibitem[Darcet et~al.(2023)Darcet, Oquab, Mairal, and Bojanowski]{darcet2023vision}
Timoth{\'e}e Darcet, Maxime Oquab, Julien Mairal, and Piotr Bojanowski.
\newblock Vision transformers need registers.
\newblock In \emph{Proceedings of the International Conference on Learning Representations (ICLR)}, 2023.

\bibitem[de~Geus et~al.(2021)de~Geus, Meletis, Lu, Wen, and Dubbelman]{de2021part}
Daan de Geus, Panagiotis Meletis, Chenyang Lu, Xiaoxiao Wen, and Gijs Dubbelman.
\newblock Part-aware panoptic segmentation.
\newblock In \emph{Proceedings of the IEEE/CVF Conference on Computer Vision and Pattern Recognition (CVPR)}, 2021.

\bibitem[Deng et~al.(2009)Deng, Dong, Socher, Li, Li, and Fei-Fei]{deng2009imagenet}
Jia Deng, Wei Dong, Richard Socher, Li-Jia Li, Kai Li, and Li Fei-Fei.
\newblock Imagenet: A large-scale hierarchical image database.
\newblock In \emph{Proceedings of the IEEE/CVF Conference on Computer Vision and Pattern Recognition (CVPR)}, 2009.

\bibitem[Duan et~al.(2024)Duan, Yang, Pan, and Liu]{duan2023lcco}
Xin Duan, Yan Yang, Liyuan Pan, and Xiabi Liu.
\newblock {LCCo}: Lending {CLIP} to co-segmentation.
\newblock \emph{Pattern Recognition}, page 111252, 2024.

\bibitem[Faktor and Irani(2013)]{faktor2013co}
Alon Faktor and Michal Irani.
\newblock Co-segmentation by composition.
\newblock In \emph{Proceedings of the IEEE/CVF International Conference on Computer Vision (ICCV)}, 2013.

\bibitem[Gao et~al.(2021)Gao, Wang, Liu, and Chen]{gao2021unsupervised}
Qingzhe Gao, Bin Wang, Libin Liu, and Baoquan Chen.
\newblock Unsupervised co-part segmentation through assembly.
\newblock In \emph{Proceedings of the International Conference on Machine Learning (ICML)}, 2021.

\bibitem[Gupta et~al.(2019)Gupta, Dollar, and Girshick]{gupta2019lvis}
Agrim Gupta, Piotr Dollar, and Ross Girshick.
\newblock {LVIS}: A dataset for large vocabulary instance segmentation.
\newblock In \emph{Proceedings of the IEEE/CVF Conference on Computer Vision and Pattern Recognition (CVPR)}, 2019.

\bibitem[He et~al.(2022)He, Yang, Yang, Kortylewski, Yuan, Chen, Liu, Yang, Yu, and Yuille]{he2022partimagenet}
Ju He, Shuo Yang, Shaokang Yang, Adam Kortylewski, Xiaoding Yuan, Jie-Neng Chen, Shuai Liu, Cheng Yang, Qihang Yu, and Alan Yuille.
\newblock {PartImageNet}: A large, high-quality dataset of parts.
\newblock In \emph{Proceedings of the European Conference on Computer Vision (ECCV)}, 2022.

\bibitem[He et~al.(2017)He, Gkioxari, Doll{\'a}r, and Girshick]{he2017mask}
Kaiming He, Georgia Gkioxari, Piotr Doll{\'a}r, and Ross Girshick.
\newblock {Mask R-CNN}.
\newblock In \emph{Proceedings of the IEEE/CVF International Conference on Computer Vision (ICCV)}, 2017.

\bibitem[Hu et~al.(2021)Hu, Wallis, Allen-Zhu, Li, Wang, Wang, Chen, et~al.]{hu2021lora}
Edward~J Hu, Phillip Wallis, Zeyuan Allen-Zhu, Yuanzhi Li, Shean Wang, Lu Wang, Weizhu Chen, et~al.
\newblock {LoRA}: Low-rank adaptation of large language models.
\newblock In \emph{Proceedings of the International Conference on Learning Representations (ICLR)}, 2021.

\bibitem[Huang et~al.(2024{\natexlab{a}})Huang, Liu, Lin, Du, Pang, and Lin]{huang2023lorahub}
Chengsong Huang, Qian Liu, Bill~Yuchen Lin, Chao Du, Tianyu Pang, and Min Lin.
\newblock {LoraHub}: Efficient cross-task generalization via dynamic lora composition.
\newblock In \emph{Conference on Language Modeling (COLM)}, 2024{\natexlab{a}}.

\bibitem[Huang et~al.(2024{\natexlab{b}})Huang, Meng, Liu, Su, Collier, and Lu]{huang2023sparkles}
Yupan Huang, Zaiqiao Meng, Fangyu Liu, Yixuan Su, Nigel Collier, and Yutong Lu.
\newblock Sparkles: Unlocking chats across multiple images for multimodal instruction-following models.
\newblock In \emph{ICLR Workshop on Navigating and Addressing Data Problems for Foundation Models}, 2024{\natexlab{b}}.

\bibitem[Hung et~al.(2019)Hung, Jampani, Liu, Molchanov, Yang, and Kautz]{hung2019scops}
Wei-Chih Hung, Varun Jampani, Sifei Liu, Pavlo Molchanov, Ming-Hsuan Yang, and Jan Kautz.
\newblock {SCOPS}: Self-supervised co-part segmentation.
\newblock In \emph{Proceedings of the IEEE/CVF Conference on Computer Vision and Pattern Recognition (CVPR)}, 2019.

\bibitem[Ji et~al.(2020)Ji, Du, Zhang, Wen, Wu, Zhao, Huang, and Lyu]{ji2020learning}
Ruyi Ji, Dawei Du, Libo Zhang, Longyin Wen, Yanjun Wu, Chen Zhao, Feiyue Huang, and Siwei Lyu.
\newblock Learning semantic neural tree for human parsing.
\newblock In \emph{Proceedings of the European Conference on Computer Vision (ECCV)}, 2020.

\bibitem[Kirillov et~al.(2019)Kirillov, He, Girshick, Rother, and Doll{\'a}r]{kirillov2019panoptic}
Alexander Kirillov, Kaiming He, Ross Girshick, Carsten Rother, and Piotr Doll{\'a}r.
\newblock Panoptic segmentation.
\newblock In \emph{Proceedings of the IEEE/CVF Conference on Computer Vision and Pattern Recognition (CVPR)}, 2019.

\bibitem[Kirillov et~al.(2023)Kirillov, Mintun, Ravi, Mao, Rolland, Gustafson, Xiao, Whitehead, Berg, Lo, et~al.]{kirillov2023segment}
Alexander Kirillov, Eric Mintun, Nikhila Ravi, Hanzi Mao, Chloe Rolland, Laura Gustafson, Tete Xiao, Spencer Whitehead, Alexander~C Berg, Wan-Yen Lo, et~al.
\newblock Segment anything.
\newblock In \emph{Proceedings of the IEEE/CVF International Conference on Computer Vision (ICCV)}, 2023.

\bibitem[Lai et~al.(2024)Lai, Tian, Chen, Li, Yuan, Liu, and Jia]{lai2024lisa}
Xin Lai, Zhuotao Tian, Yukang Chen, Yanwei Li, Yuhui Yuan, Shu Liu, and Jiaya Jia.
\newblock {LISA}: Reasoning segmentation via large language model.
\newblock In \emph{Proceedings of the IEEE/CVF Conference on Computer Vision and Pattern Recognition (CVPR)}, 2024.

\bibitem[Li et~al.(2019)Li, Sun, Li, Wu, and Hu]{li2019group}
Bo Li, Zhengxing Sun, Qian Li, Yunjie Wu, and Anqi Hu.
\newblock Group-wise deep object co-segmentation with co-attention recurrent neural network.
\newblock In \emph{Proceedings of the IEEE/CVF International Conference on Computer Vision (ICCV)}, 2019.

\bibitem[Li et~al.(2024{\natexlab{a}})Li, Zhang, Sun, Zou, Liu, Yang, Li, Zhang, and Gao]{li2023semantic}
Feng Li, Hao Zhang, Peize Sun, Xueyan Zou, Shilong Liu, Jianwei Yang, Chunyuan Li, Lei Zhang, and Jianfeng Gao.
\newblock {Semantic-SAM}: Segment and recognize anything at any granularity.
\newblock In \emph{Proceedings of the European Conference on Computer Vision (ECCV)}, 2024{\natexlab{a}}.

\bibitem[Li et~al.(2023{\natexlab{a}})Li, Li, Savarese, and Hoi]{li2023blip}
Junnan Li, Dongxu Li, Silvio Savarese, and Steven Hoi.
\newblock {BLIP-2}: Bootstrapping language-image pre-training with frozen image encoders and large language models.
\newblock In \emph{International Conference on Machine Learning (ICML)}, 2023{\natexlab{a}}.

\bibitem[Li et~al.(2023{\natexlab{b}})Li, Pan, Ge, Gao, Ji, Zhang, Chua, Tang, Zhang, and Zhuang]{li2023fine}
Juncheng Li, Kaihang Pan, Zhiqi Ge, Minghe Gao, Wei Ji, Wenqiao Zhang, Tat-Seng Chua, Siliang Tang, Hanwang Zhang, and Yueting Zhuang.
\newblock Fine-tuning multimodal llms to follow zero-shot demonstrative instructions.
\newblock In \emph{Proceedings of the International Conference on Learning Representations (ICLR)}, 2023{\natexlab{b}}.

\bibitem[Li et~al.(2024{\natexlab{b}})Li, Wu, Zhao, Bai, and Bai]{li2024partglee}
Junyi Li, Junfeng Wu, Weizhi Zhao, Song Bai, and Xiang Bai.
\newblock {PartGLEE}: A foundation model for recognizing and parsing any objects.
\newblock In \emph{Proceedings of the European Conference on Computer Vision (ECCV)}, 2024{\natexlab{b}}.

\bibitem[Li et~al.(2022)Li, Xu, Yang, Cheng, Tong, and Tao]{li2022panoptic}
Xiangtai Li, Shilin Xu, Yibo Yang, Guangliang Cheng, Yunhai Tong, and Dacheng Tao.
\newblock {Panoptic-PartFormer}: Learning a unified model for panoptic part segmentation.
\newblock In \emph{Proceedings of the European Conference on Computer Vision (ECCV)}, 2022.

\bibitem[Lin et~al.(2014)Lin, Maire, Belongie, Hays, Perona, Ramanan, Doll{\'a}r, and Zitnick]{lin2014microsoft}
Tsung-Yi Lin, Michael Maire, Serge Belongie, James Hays, Pietro Perona, Deva Ramanan, Piotr Doll{\'a}r, and C~Lawrence Zitnick.
\newblock Microsoft {COCO}: Common objects in context.
\newblock In \emph{Proceedings of the European Conference on Computer Vision (ECCV)}, 2014.

\bibitem[Lin et~al.(2017)Lin, Goyal, Girshick, He, and Dollár]{ross2017focal}
Tsung-Yi Lin, Priya Goyal, Ross Girshick, Kaiming He, and Piotr Dollár.
\newblock Focal loss for dense object detection.
\newblock In \emph{Proceedings of the IEEE/CVF International Conference on Computer Vision (ICCV)}, 2017.

\bibitem[Liu et~al.(2024)Liu, Li, Wu, and Lee]{liu2024visual}
Haotian Liu, Chunyuan Li, Qingyang Wu, and Yong~Jae Lee.
\newblock Visual instruction tuning.
\newblock \emph{Proceedings of the Conference on Neural Information Processing Systems (NeurIPS)}, 36, 2024.

\bibitem[Liu et~al.(2021{\natexlab{a}})Liu, Zhang, Yang, Su, and Zhu]{liu2021unsupervised}
Shilong Liu, Lei Zhang, Xiao Yang, Hang Su, and Jun Zhu.
\newblock Unsupervised part segmentation through disentangling appearance and shape.
\newblock In \emph{Proceedings of the IEEE/CVF Conference on Computer Vision and Pattern Recognition (CVPR)}, 2021{\natexlab{a}}.

\bibitem[Liu et~al.(2021{\natexlab{b}})Liu, Lin, Cao, Hu, Wei, Zhang, Lin, and Guo]{liu2021swin}
Ze Liu, Yutong Lin, Yue Cao, Han Hu, Yixuan Wei, Zheng Zhang, Stephen Lin, and Baining Guo.
\newblock Swin transformer: Hierarchical vision transformer using shifted windows.
\newblock In \emph{Proceedings of the IEEE/CVF International Conference on Computer Vision (ICCV)}, 2021{\natexlab{b}}.

\bibitem[Long et~al.(2015)Long, Shelhamer, and Darrell]{long2015fully}
Jonathan Long, Evan Shelhamer, and Trevor Darrell.
\newblock Fully convolutional networks for semantic segmentation.
\newblock In \emph{Proceedings of the IEEE/CVF Conference on Computer Vision and Pattern Recognition (CVPR)}, 2015.

\bibitem[Loshchilov and Hutter(2018)]{loshchilov2018decoupled}
Ilya Loshchilov and Frank Hutter.
\newblock Decoupled weight decay regularization.
\newblock In \emph{Proceedings of the International Conference on Learning Representations (ICLR)}, 2018.

\bibitem[Lu et~al.(2021)Lu, Qiu, Chen, Xia, Zhao, Zhang, Yu, Liang, and Zhu]{lu2021iconqa}
Pan Lu, Liang Qiu, Jiaqi Chen, Tony Xia, Yizhou Zhao, Wei Zhang, Zhou Yu, Xiaodan Liang, and Song-Chun Zhu.
\newblock {IconQA}: A new benchmark for abstract diagram understanding and visual language reasoning.
\newblock In \emph{Proceedings of the Conference on Neural Information Processing Systems (NeurIPS) Datasets and Benchmarks Track}, 2021.

\bibitem[Michieli and Zanuttigh(2022)]{michieli2022edge}
Umberto Michieli and Pietro Zanuttigh.
\newblock Edge-aware graph matching network for part-based semantic segmentation.
\newblock \emph{International Journal of Computer Vision (IJCV)}, 130\penalty0 (11):\penalty0 2797--2821, 2022.

\bibitem[Michieli et~al.(2020)Michieli, Borsato, Rossi, and Zanuttigh]{michieli2020gmnet}
Umberto Michieli, Edoardo Borsato, Luca Rossi, and Pietro Zanuttigh.
\newblock {GMNet}: Graph matching network for large scale part semantic segmentation in the wild.
\newblock In \emph{Proceedings of the European Conference on Computer Vision (ECCV)}, 2020.

\bibitem[Milletari et~al.(2016)Milletari, Navab, and Ahmadi]{milletari2016v}
Fausto Milletari, Nassir Navab, and Seyed-Ahmad Ahmadi.
\newblock {V-Net}: Fully convolutional neural networks for volumetric medical image segmentation.
\newblock In \emph{Proceedings of the International Conference on 3D Vision (3DV)}, 2016.

\bibitem[Oquab et~al.(2023)Oquab, Darcet, Moutakanni, Vo, Szafraniec, Khalidov, Fernandez, HAZIZA, Massa, El-Nouby, et~al.]{oquab2023dinov2}
Maxime Oquab, Timoth{\'e}e Darcet, Th{\'e}o Moutakanni, Huy~V Vo, Marc Szafraniec, Vasil Khalidov, Pierre Fernandez, Daniel HAZIZA, Francisco Massa, Alaaeldin El-Nouby, et~al.
\newblock {DINOv2}: Learning robust visual features without supervision.
\newblock \emph{Transactions on Machine Learning Research}, 2023.

\bibitem[Paszke et~al.(2019)Paszke, Gross, Massa, Lerer, Bradbury, Chanan, Killeen, Lin, Gimelshein, Antiga, et~al.]{paszke2019pytorch}
Adam Paszke, Sam Gross, Francisco Massa, Adam Lerer, James Bradbury, Gregory Chanan, Trevor Killeen, Zeming Lin, Natalia Gimelshein, Luca Antiga, et~al.
\newblock {PyTorch}: An imperative style, high-performance deep learning library.
\newblock In \emph{Proceedings of the Conference in Neural Information Processing Systems (NeurIPS)}, 2019.

\bibitem[Pramanick et~al.(2024)Pramanick, Han, Hou, Nag, Lim, Ballas, Wang, Chellappa, and Almahairi]{pramanick2024jack}
Shraman Pramanick, Guangxing Han, Rui Hou, Sayan Nag, Ser-Nam Lim, Nicolas Ballas, Qifan Wang, Rama Chellappa, and Amjad Almahairi.
\newblock Jack of all tasks master of many: Designing general-purpose coarse-to-fine vision-language model.
\newblock In \emph{Proceedings of the IEEE/CVF Conference on Computer Vision and Pattern Recognition (CVPR)}, 2024.

\bibitem[Radford et~al.(2021)Radford, Kim, Hallacy, Ramesh, Goh, Agarwal, Sastry, Askell, Mishkin, Clark, et~al.]{radford2021learning}
Alec Radford, Jong~Wook Kim, Chris Hallacy, Aditya Ramesh, Gabriel Goh, Sandhini Agarwal, Girish Sastry, Amanda Askell, Pamela Mishkin, Jack Clark, et~al.
\newblock Learning transferable visual models from natural language supervision.
\newblock In \emph{Proceedings of the International Conference on Machine Learning (ICML)}, 2021.

\bibitem[Ramanathan et~al.(2023)Ramanathan, Kalia, Petrovic, Wen, Zheng, Guo, Wang, Marquez, Kovvuri, Kadian, et~al.]{ramanathan2023paco}
Vignesh Ramanathan, Anmol Kalia, Vladan Petrovic, Yi Wen, Baixue Zheng, Baishan Guo, Rui Wang, Aaron Marquez, Rama Kovvuri, Abhishek Kadian, et~al.
\newblock {PACO}: Parts and attributes of common objects.
\newblock In \emph{Proceedings of the IEEE/CVF Conference on Computer Vision and Pattern Recognition (CVPR)}, 2023.

\bibitem[Rasheed et~al.(2024)Rasheed, Maaz, Shaji, Shaker, Khan, Cholakkal, Anwer, Xing, Yang, and Khan]{rasheed2023glamm}
Hanoona Rasheed, Muhammad Maaz, Sahal Shaji, Abdelrahman Shaker, Salman Khan, Hisham Cholakkal, Rao~M Anwer, Erix Xing, Ming-Hsuan Yang, and Fahad~S Khan.
\newblock {GLaMM}: Pixel grounding large multimodal model.
\newblock In \emph{Proceedings of the IEEE/CVF Conference on Computer Vision and Pattern Recognition (CVPR)}, 2024.

\bibitem[Rasley et~al.(2020)Rasley, Rajbhandari, Ruwase, and He]{rasley2020deepspeed}
Jeff Rasley, Samyam Rajbhandari, Olatunji Ruwase, and Yuxiong He.
\newblock {DeepSpeed}: System optimizations enable training deep learning models with over 100 billion parameters.
\newblock In \emph{Proceedings of the ACM SIGKDD International Conference on Knowledge Discovery \& Data Mining}, 2020.

\bibitem[Reimers and Gurevych(2019)]{reimers2019sentence}
Nils Reimers and Iryna Gurevych.
\newblock Sentence-{BERT}: Sentence embeddings using {S}iamese {BERT}-networks.
\newblock In \emph{Proceedings of the Conference on Empirical Methods in Natural Language Processing and the International Joint Conference on Natural Language Processing (EMNLP-IJCNLP)}, 2019.

\bibitem[Ren et~al.(2024{\natexlab{a}})Ren, Li, Wang, Zhao, and Qin]{ren2024analyzing}
Weijieying Ren, Xinlong Li, Lei Wang, Tianxiang Zhao, and Wei Qin.
\newblock Analyzing and reducing catastrophic forgetting in parameter efficient tuning.
\newblock \emph{arXiv preprint arXiv:2402.18865}, 2024{\natexlab{a}}.

\bibitem[Ren et~al.(2024{\natexlab{b}})Ren, Huang, Wei, Zhao, Fu, Feng, and Jin]{ren2023pixellm}
Zhongwei Ren, Zhicheng Huang, Yunchao Wei, Yao Zhao, Dongmei Fu, Jiashi Feng, and Xiaojie Jin.
\newblock {PixelLM}: Pixel reasoning with large multimodal model.
\newblock In \emph{Proceedings of the IEEE/CVF Conference on Computer Vision and Pattern Recognition (CVPR)}, 2024{\natexlab{b}}.

\bibitem[Rubinstein et~al.(2013)Rubinstein, Joulin, Kopf, and Liu]{rubinstein2013unsupervised}
Michael Rubinstein, Armand Joulin, Johannes Kopf, and Ce Liu.
\newblock Unsupervised joint object discovery and segmentation in internet images.
\newblock In \emph{Proceedings of the IEEE/CVF Conference on Computer Vision and Pattern Recognition (CVPR)}, 2013.

\bibitem[Shotton et~al.(2006)Shotton, Winn, Rother, and Criminisi]{shotton2006textonboost}
Jamie Shotton, John Winn, Carsten Rother, and Antonio Criminisi.
\newblock {TextonBoost}: Joint appearance, shape and context modeling for multi-class object recognition and segmentation.
\newblock In \emph{Proceedings of the European Conference on Computer Vision (ECCV)}, 2006.

\bibitem[Singh et~al.(2022)Singh, Gupta, Shenoy, and Sarvadevabhatla]{singh2022float}
Rishubh Singh, Pranav Gupta, Pradeep Shenoy, and Ravikiran Sarvadevabhatla.
\newblock {FLOAT}: Factorized learning of object attributes for improved multi-object multi-part scene parsing.
\newblock In \emph{Proceedings of the IEEE/CVF Conference on Computer Vision and Pattern Recognition (CVPR)}, 2022.

\bibitem[Suhr et~al.(2017)Suhr, Lewis, Yeh, and Artzi]{suhr2017corpus}
Alane Suhr, Mike Lewis, James Yeh, and Yoav Artzi.
\newblock A corpus of natural language for visual reasoning.
\newblock In \emph{Proceedings of the Annual Meeting of the Association for Computational Linguistics (ACL)}, 2017.

\bibitem[Suhr et~al.(2019)Suhr, Zhou, Zhang, Zhang, Bai, and Artzi]{suhr2018corpus}
Alane Suhr, Stephanie Zhou, Ally Zhang, Iris Zhang, Huajun Bai, and Yoav Artzi.
\newblock A corpus for reasoning about natural language grounded in photographs.
\newblock In \emph{Proceedings of the Annual Meeting of the Association for Computational Linguistics (ACL)}, 2019.

\bibitem[Sun et~al.(2023{\natexlab{a}})Sun, Chen, Zhu, Xiao, Luo, Xie, and Yan]{sun2023going}
Peize Sun, Shoufa Chen, Chenchen Zhu, Fanyi Xiao, Ping Luo, Saining Xie, and Zhicheng Yan.
\newblock Going denser with open-vocabulary part segmentation.
\newblock In \emph{Proceedings of the IEEE/CVF International Conference on Computer Vision (ICCV)}, 2023{\natexlab{a}}.

\bibitem[Sun et~al.(2023{\natexlab{b}})Sun, Fang, Wu, Wang, and Cao]{sun2023eva}
Quan Sun, Yuxin Fang, Ledell Wu, Xinlong Wang, and Yue Cao.
\newblock {Eva-CLIP}: Improved training techniques for clip at scale.
\newblock \emph{arXiv preprint arXiv:2303.15389}, 2023{\natexlab{b}}.

\bibitem[van~der Klis et~al.(2023)van~der Klis, Alaniz, Mancini, Dantas, Ienco, Akata, and Marcos]{van2023pdisconet}
Robert van~der Klis, Stephan Alaniz, Massimiliano Mancini, Cassio~F Dantas, Dino Ienco, Zeynep Akata, and Diego Marcos.
\newblock {PDiscoNet}: Semantically consistent part discovery for fine-grained recognition.
\newblock In \emph{Proceedings of the IEEE/CVF International Conference on Computer Vision (ICCV)}, 2023.

\bibitem[Wang et~al.(2023)Wang, Chen, Ge, Xia, Bao, Zheng, Zhang, Gui, and Huang]{wang2023orthogonal}
Xiao Wang, Tianze Chen, Qiming Ge, Han Xia, Rong Bao, Rui Zheng, Qi Zhang, Tao Gui, and Xuan-Jing Huang.
\newblock Orthogonal subspace learning for language model continual learning.
\newblock In \emph{Findings of the Conference on Empirical Methods in Natural Language Processing (EMNLP)}, 2023.

\bibitem[Wei et~al.(2024{\natexlab{a}})Wei, Tan, Zhong, Yang, and Ma]{wei2024lasagna}
Cong Wei, Haoxian Tan, Yujie Zhong, Yujiu Yang, and Lin Ma.
\newblock {LaSagnA}: Language-based segmentation assistant for complex queries.
\newblock \emph{arXiv preprint arXiv:2404.08506}, 2024{\natexlab{a}}.

\bibitem[Wei et~al.(2024{\natexlab{b}})Wei, Yue, Zhang, Kong, Liu, and Pang]{wei2024ov}
Meng Wei, Xiaoyu Yue, Wenwei Zhang, Shu Kong, Xihui Liu, and Jiangmiao Pang.
\newblock {OV-PARTS}: Towards open-vocabulary part segmentation.
\newblock In \emph{Proceedings of the Conference on Neural Information Processing Systems (NeurIPS)}, 2024{\natexlab{b}}.

\bibitem[Welinder et~al.(2010)Welinder, Branson, Mita, Wah, Schroff, Belongie, and Perona]{welinder2010caltech}
Peter Welinder, Steve Branson, Takeshi Mita, Catherine Wah, Florian Schroff, Serge Belongie, and Pietro Perona.
\newblock {Caltech-UCSD birds 200}.
\newblock 2010.

\bibitem[Xiao et~al.(2018)Xiao, Liu, Zhou, Jiang, and Sun]{xiao2018unified}
Tete Xiao, Yingcheng Liu, Bolei Zhou, Yuning Jiang, and Jian Sun.
\newblock Unified perceptual parsing for scene understanding.
\newblock In \emph{Proceedings of the European Conference on Computer Vision (ECCV)}, 2018.

\bibitem[Xin et~al.(2020)Xin, Tang, Lee, Yu, and Lin]{xin2020deebert}
Ji Xin, Raphael Tang, Jaejun Lee, Yaoliang Yu, and Jimmy Lin.
\newblock {DeeBERT}: Dynamic early exiting for accelerating bert inference.
\newblock In \emph{Proceedings of the Annual Meeting of the Association for Computational Linguistics (ACL)}, pages 2246--2251, 2020.

\bibitem[Yang et~al.(2019)Yang, Song, Wang, and Jiang]{yang2019parsing}
Lu Yang, Qing Song, Zhihui Wang, and Ming Jiang.
\newblock Parsing {R-CNN} for instance-level human analysis.
\newblock In \emph{Proceedings of the IEEE/CVF Conference on Computer Vision and Pattern Recognition (CVPR)}, 2019.

\bibitem[Yang et~al.(2020)Yang, Song, Wang, Hu, Liu, Xin, Jia, and Xu]{yang2020renovating}
Lu Yang, Qing Song, Zhihui Wang, Mengjie Hu, Chun Liu, Xueshi Xin, Wenhe Jia, and Songcen Xu.
\newblock Renovating parsing {R-CNN} for accurate multiple human parsing.
\newblock In \emph{Proceedings of the European Conference on Computer Vision (ECCV)}, 2020.

\bibitem[Yuan et~al.(2024)Yuan, Li, Liu, Tang, Luo, Qin, Zhang, and Zhu]{yuan2023osprey}
Yuqian Yuan, Wentong Li, Jian Liu, Dongqi Tang, Xinjie Luo, Chi Qin, Lei Zhang, and Jianke Zhu.
\newblock Osprey: Pixel understanding with visual instruction tuning.
\newblock In \emph{Proceedings of the IEEE/CVF International Conference on Computer Vision (CVPR)}, 2024.

\bibitem[Zhang et~al.(2024{\natexlab{a}})Zhang, Yao, Ji, Liu, and Chua]{zhang2023next}
Ao Zhang, Yuan Yao, Wei Ji, Zhiyuan Liu, and Tat-Seng Chua.
\newblock {NExT-Chat}: An lmm for chat, detection and segmentation.
\newblock In \emph{Proceedings of the International Conference on Machine Learning (ICML)}, 2024{\natexlab{a}}.

\bibitem[Zhang et~al.(2021)Zhang, Li, Lin, Wu, and Yao]{zhang2021cyclesegnet}
Chi Zhang, Guankai Li, Guosheng Lin, Qingyao Wu, and Rui Yao.
\newblock {CycleSegNet}: Object co-segmentation with cycle refinement and region correspondence.
\newblock \emph{IEEE Transactions on Image Processing}, 30:\penalty0 5652--5664, 2021.

\bibitem[Zhang et~al.(2023)Zhang, Liu, He, et~al.]{zhang2023composing}
Jinghan Zhang, Junteng Liu, Junxian He, et~al.
\newblock Composing parameter-efficient modules with arithmetic operation.
\newblock In \emph{Proceedings of the Conference on Neural Information Processing Systems (NeurIPS)}, 2023.

\bibitem[Zhang et~al.(2024{\natexlab{b}})Zhang, Herrmann, Hur, Polania~Cabrera, Jampani, Sun, and Yang]{zhang2024tale}
Junyi Zhang, Charles Herrmann, Junhwa Hur, Luisa Polania~Cabrera, Varun Jampani, Deqing Sun, and Ming-Hsuan Yang.
\newblock A tale of two features: Stable diffusion complements dino for zero-shot semantic correspondence.
\newblock In \emph{Proceedings of the Conference on Neural Information Processing Systems (NeurIPS)}, 2024{\natexlab{b}}.

\bibitem[Zhang et~al.(2020)Zhang, Chen, Liu, and Liu]{zhang2020deep}
Kaihua Zhang, Jin Chen, Bo Liu, and Qingshan Liu.
\newblock Deep object co-segmentation via spatial-semantic network modulation.
\newblock In \emph{Proceedings of the AAAI Conference on Artificial Intelligence}, 2020.

\bibitem[Zhou et~al.(2017)Zhou, Zhao, Puig, Fidler, Barriuso, and Torralba]{zhou2017scene}
Bolei Zhou, Hang Zhao, Xavier Puig, Sanja Fidler, Adela Barriuso, and Antonio Torralba.
\newblock Scene parsing through {ADE20K} dataset.
\newblock In \emph{Proceedings of the IEEE/CVF Conference on Computer Vision and Pattern Recognition (CVPR)}, 2017.

\bibitem[Zhou et~al.(2019)Zhou, Zhao, Puig, Xiao, Fidler, Barriuso, and Torralba]{zhou2019semantic}
Bolei Zhou, Hang Zhao, Xavier Puig, Tete Xiao, Sanja Fidler, Adela Barriuso, and Antonio Torralba.
\newblock Semantic understanding of scenes through the {ADE20K} dataset.
\newblock \emph{International Journal of Computer Vision (IJCV)}, 127:\penalty0 302--321, 2019.

\bibitem[Zhou et~al.(2021)Zhou, Wang, Liu, Yang, and Van~Gool]{zhou2021differentiable}
Tianfei Zhou, Wenguan Wang, Si Liu, Yi Yang, and Luc Van~Gool.
\newblock Differentiable multi-granularity human representation learning for instance-aware human semantic parsing.
\newblock In \emph{Proceedings of the IEEE/CVF Conference on Computer Vision and Pattern Recognition (CVPR)}, 2021.

\bibitem[Zhou et~al.(2023)Zhou, Yang, and Wang]{zhou2023differentiable}
Tianfei Zhou, Yi Yang, and Wenguan Wang.
\newblock Differentiable multi-granularity human parsing.
\newblock \emph{IEEE Transactions on Pattern Analysis and Machine Intelligence (TPAMI)}, 2023.

\bibitem[Zou et~al.(2024)Zou, Yang, Zhang, Li, Li, Wang, Wang, Gao, and Lee]{zou2024segment}
Xueyan Zou, Jianwei Yang, Hao Zhang, Feng Li, Linjie Li, Jianfeng Wang, Lijuan Wang, Jianfeng Gao, and Yong~Jae Lee.
\newblock Segment everything everywhere all at once.
\newblock In \emph{Proceedings of the Conference on Neural Information Processing Systems (NeurIPS)}, 2024.

\end{thebibliography}
}

\appendix
\clearpage
\setcounter{page}{1}
\maketitlesupplementary

\section{\benchmarkname{}}\label{appendix:mixedparts}

\subsection{Single-Image Part Segmentation Datasets}
\label{sec:public-datasets}
To construct a robust dataset for part-focused semantic co-segmentation, we carefully curate generalizable and diverse data at various levels of detail. This entails selecting datasets that cover a wide range of objects and parts, both rigid (\eg, utensils, vehicles, scissors) and non-rigid (\eg, animals, humans), while not being too domain-specific (\eg, birds or celebrities' faces), which could potentially encourage overfitting or introduce bias in training. Therefore, we select the following datasets to construct \benchmarkname{}:
\begin{itemize}[label=\faPaw, topsep=0cm, partopsep=0pt, parsep=0pt, itemsep=0pt]
    \item \textbf{PartImageNet} \cite{he2022partimagenet} is a high-quality part-focused extension of ImageNet \cite{deng2009imagenet} covering a variety of object classes with mostly animals (\eg bird, fish, \etc) to facilitate non-rigid part understanding. Each image in PartImageNet contains only one foreground object, which can encourage the model to focus on important foreground objects when comparing two images. Prior to use, we make modifications to the original dataset to enhance generalizability, as detailed in Appendix \ref{sec:partimagenet}.
    \item \textbf{ADE20K-Part234} \cite{wei2024ov} is a revised version of the ADE20K scene parsing dataset \cite{zhou2017scene, zhou2019semantic} with an emphasis on object parts. The original ADE20K dataset encompasses a wide variety of scenes, including indoor, outdoor, and urban environments, which naturally lends itself to more complex visual signals covering multiple objects in contrast to PartImageNet. However, less than $15\%$ of the dataset contains part annotations; in addition, some part labels are too granular, which may encourage overfitting while not being too beneficial for general part understanding (\eg ``table stretcher'' and ``table h-stretcher''). To amend these disadvantages, \citet{wei2024ov} introduces ADE20K-Part234, a clean, part-focused version of ADE20K for improved part analysis. 
    \item \textbf{PACO-LVIS} \cite{ramanathan2023paco} is a part-centric version of LVIS \cite{gupta2019lvis} that is based on COCO \cite{lin2014microsoft} and focuses on diverse everyday objects, further contributing to the diversity of object categories in \benchmarkname{}. 
    PACO contains an extensive list of object-part categories as well as complex images with multiple objects and parts, providing finer granularity for part understanding. 
\end{itemize}

To curate \benchmarkname{}, we first select 1,885 pairs of object categories across all 3 datasets that have at least one common part label (\eg ``armchair's seat'' and ``swivel chair's seat'') to ensure annotation availability. However, due to the ambiguity of natural language, the same part name can refer to object parts that are not typically intuitively comparable. For example, even though both a bus and a microwave oven may have a door, they are not commonly compared. Therefore, we manually curate intuitively comparable object pairs from all possible pairs, resulting in 964 pairs of categories across all 3 datasets.

With the object pairings available, we pair up individual images corresponding to our common object, common part, and unique part localization subtasks. For common object parts, we select images that have at least a common visible object and/or object part. However, since different object classes can share the same parts, we also include images of logically comparable objects of different classes, for instance, a chair and an ottoman (both seating furniture) or an airplane and a bird (both having wings and usually compared as flying objects). We ensure stratification of object categories in \benchmarkname{} to reflect the data distribution in the original datasets.

\subsection{PartImageNet Modifications}
\label{sec:partimagenet}

While PartImageNet provides high-quality segmentation masks across a diverse range of object classes, its categories are often too abstract and not commonly used in natural language due to their broad scope. For instance, PartImageNet classifies all four-legged animals under the ``quadruped'' supercategory. Although technically accurate, such classifications are too generic for practical use in everyday language.

To address this, we manually selected the most commonly used object class associated with each category provided by ImageNet.
The dataset includes WordNet synset IDs (\eg, ``n02071294'') that correspond to specific object classes. 
Using these synset IDs, we extracted the hierarchical path of each class within the WordNet graph. For example, the synset ID ``n02071294'' maps to the following WordNet path: living\_thing $\to$ organism $\to$ animal $\to$ chordate $\to$ vertebrate $\to$ mammal $\to$ placental $\to$ aquatic\_mammal $\to$ cetacean $\to$ whale $\to$ toothed\_whale $\to$ dolphin $\to$ killer\_whale. From this path, we selected \textit{whale} as the representative object category for this class. We repeat this process for all object classes provided by PartImageNet. The object classes we use alongside the original supercategory in parentheses are shown in Table \ref{tab:obj-parts2}.

\section{\benchmarkname{} Dataset Statistics}
\label{appendix:datasets}
\begin{figure}[t!]
  \centering
  \includegraphics[width=0.9\linewidth]{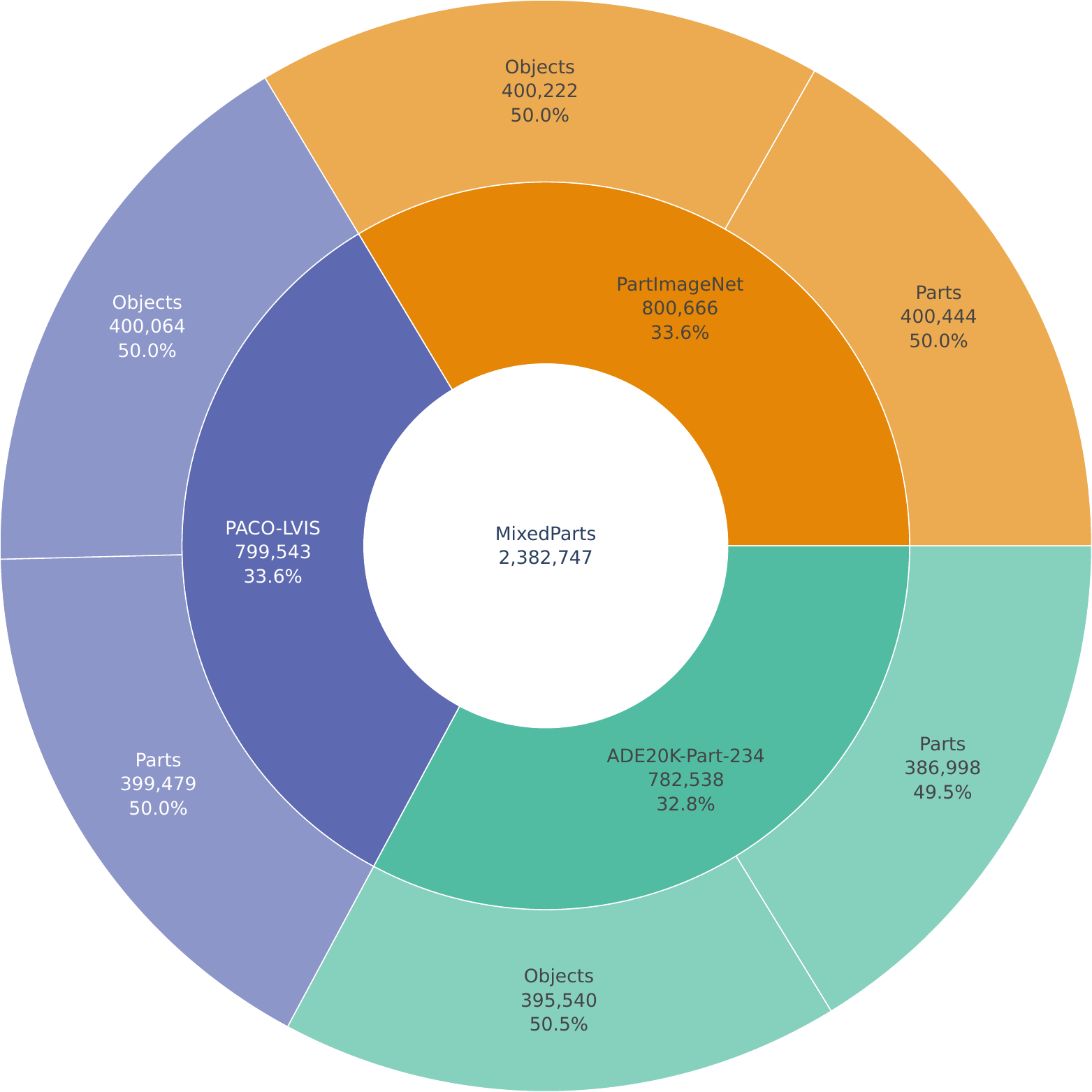}
  \caption{\textbf{\benchmarkname{} Dataset Overview.}}
  \label{fig:dataset-overview}
  \vspace{-0.4cm}
\end{figure}

Figure \ref{fig:dataset-overview} and Table \ref{tab:pairs-stats} provide a comprehensive overview of our benchmark, illustrating its structure and the distribution of objects and parts. The inner circle of the donut chart represents the entire dataset, comprising a total of 2,382,747 samples (image pairs), sourced from PartImageNet, PACO-LVIS, and ADE20K-Part234, each contributing to roughly a third of the total dataset. The outer ring of the chart further divides these sources into categories of objects and parts,  contributing roughly half, ensuring balanced representation of various object–part relationships.
In \benchmarkname{}, there are 2,509,552 instances of common objects, 3,443,758 instances of common parts, and 5,557,188 instances of unique parts. 
On average, each image pair includes 2 object instances (one per image), 4 common part instances, and 5.3 unique part instances. Sample complexity varies widely, with a maximum of 10 common objects, 56 common parts, and 72 unique parts per sample. 
Median counts are 2 for common objects, 4 for common parts, and 3 for unique parts, reflecting that a single object pair can involve multiple shared and unique parts.
The dataset spans 141 object categories and 196 part categories, covering a wide variety of objects and parts, totaling 65,960 distinct object instances and 154,463 distinct part instances. 
Figures \ref{fig:object-distribution}-\ref{fig:part-distribution} present the top-30 most frequent object and part categories, while Figures \ref{fig:common-part-distribution}-\ref{fig:unique-part-distribution} show the top-30 common and unique parts, respectively.

\section{Additional Experimental Setup Details}
\label{appendix:impl}
\subsection{Implementation Details}
\paragraph{\modelname{}.} We use PyTorch~\cite{paszke2019pytorch} to implement and optimize \modelname{}, with DeepSpeed~\cite{rasley2020deepspeed} for efficient training. We initialize our model on GLaMM's \cite{rasheed2023glamm} pretrained checkpoint,\footnote{\scriptsize \url{https://huggingface.co/MBZUAI/GLaMM-GranD-Pretrained}.} the Q-Former module on InstructBLIP's \cite{dai2023instructblip} Vicuna-7B-aligned checkpoint,\footnote{\scriptsize \url{https://storage.googleapis.com/sfr-vision-language-research/LAVIS/models/InstructBLIP/instruct_blip_vicuna7b_trimmed.pth}.} and align them with 3.2M samples from GLaMM's GranD dataset. All training is conducted on four NVIDIA A40 GPUs with 48GB memory. LoRA layers are applied to the query and value projections, configured with a rank of 8 and a scaling factor of 16. We train for 10 epochs, each comprising 500 steps, using the AdamW optimizer~\cite{loshchilov2018decoupled} with a 4e-4 initial learning rate and beta coefficients set to 0.9 and 0.95, respectively. We warm up the learning rate over 100 training steps and then use a linear decay scheduler over the remaining training time. We employ a dropout rate of 0.05, 1.0 gradient clipping, and set the batch size to 4 alongside gradient accumulation every 4 steps. The loss coefficients are set to $\lambda_{\text{text}}\!=\!1.0$, $\lambda_{\text{focal}}\!=\!2.0$, and $\lambda_{\text{Dice}}\!=\!0.5$. We retain all original implementation details for our baselines while keeping the effective batch size the same across all finetuned approaches to ensure fairness. We use DeepSpeed's profiler\footnote{\scriptsize \url{https://www.deepspeed.ai/tutorials/flops-profiler/}.} to compute TFLOPS.
\vspace{0.1cm}

\begin{table*}[t!]
\centering
\begin{tabular}{l|ccc} \toprule
\textbf{Statistic} & \textbf{Common Objects} & \textbf{Common Parts} & \textbf{Unique Parts} \\ \hline
Total \# Instances & 2,509,552 & 3,443,758 & 5,557,188 \\
Average \# Instances/Sample & 2.099 & 4.037 & 5.269 \\
Maximum \# Instances/Sample & 10 & 56 & 72 \\
Median \# Instances/Sample & 2 & 4 & 3 \\ \bottomrule
\end{tabular}
\vspace{-0.2cm}
\caption{\textbf{\benchmarkname{} Statistics} on \# instance of objects/parts. Here, an object/part instance in \benchmarkname{} refers to a single occurrence of an object or part in a sample, while a sample corresponds to an image pair.}
\label{tab:pairs-stats}
\vspace{-0.4cm}
\end{table*}

\noindent \textbf{Baselines.} We provide implementation details for our baselines, which include three zero-shot and two finetuned LVLM-based approaches:
\begin{itemize}[label=\faPaw, topsep=0cm, partopsep=0pt, parsep=0pt, itemsep=0pt]
\item \textbf{Cascade:} To evaluate whether a simple modular pipeline can effectively solve part-focused semantic co-segmentation, we first design an cascaded approach that sequentially integrates strong zero-shot models, each excelling at different subcomponents of the task: (1) Sparkles \cite{huang2023sparkles}, a powerful multi-image model, is used to identify relevant objects across images, followed by (2) GPT-4o API \cite{achiam2023gpt} to extract object-part relationships, and finally, (3) LISA \cite{lai2024lisa}, a segmentation-based LVLM, to produce segmentation masks for the identified objects and parts. Given an image pair, we first prompt Sparkles to list the objects or parts present across both images. Next, we prompt GPT-4o with this information to identify the common or unique objects or parts, depending on the task. Finally, we prompt LISA with the objects/parts extracted by GPT-4o to generate pixel-level segmentation masks.
Prompt templates for all models of this cascaded approach are shown in Figure \ref{fig:cascade-prompts}.
\item \textbf{Multi-Image VLPart:} VLPart \cite{sun2023going} is an open-vocabulary part segmentation model that can segment object parts at different granularities. It combines a conventional Mask R-CNN \cite{he2017mask} with a modern Swin Transformer backbone \cite{liu2021swin} and a CLIP \cite{radford2021learning} text classifier for open-world image classification. Although VLPart cannot perform co-segmentation, its strong zero-shot object-part segmentation capabilities enable a simple baseline: performing segmentation on individual images and simply examining the common and unique predictions. 
Specifically, for a given image pair, we perform inference on both images and extract all predicted object and part masks. To identify common objects, we compare the sets of predicted objects across images. For common and unique parts, we aggregate the sets of all predicted parts and derive their associated objects. We follow the default configuration from the official repository,\footnote{\scriptsize \url{https://github.com/facebookresearch/VLPart}.} including the detection confidence threshold of 0.7 (\ie, retaining predictions with confidence scores above 0.7) to initialize the model. We use the checkpoint with a Swin-base cascade backbone trained on the parsed ImageNet dataset.\footnote{\scriptsize \url{https://github.com/PeizeSun/VLPart/releases/download/v0.1/swinbase_cascade_lvis_paco_pascalpart_partimagenet_inparsed.pth}.}
\item \textbf{Multi-Image PartGLEE:} PartGLEE \cite{li2024partglee} is a recent foundation model that uses Q-Former to query object and part representations from large-scale training data, boasting better object-part segmentation capabilities than VLPart. For this baseline, we simply obtain single-image outputs from PartGLEE (100 masks per image), keeping only those with non-zero scores (55 on average). Similar to Multi-Image VLPart, we compare the predicted class labels across the image pair to identify common and unique objects and parts. We use PartGLEE's official implementation,\footnote{\scriptsize \url{https://github.com/ProvenceStar/PartGLEE.git}.} initializing with the released Swin-Large checkpoint and default settings.
\item \textbf{Multi-Image LISA:} LISA \cite{lai2024lisa} is an LVLM trained to perform referring segmentation, built on a Vicuna-7B backbone and SAM. Since LISA was not trained for multi-image processing, we replicate \modelname{}'s multi-image implementation on the LISA codebase. We initialize LISA on their 7B checkpoint\footnote{\scriptsize \url{https://huggingface.co/xinlai/LISA-7B-v1}.} and finetune the mask decoder alongside LoRA.
\item \textbf{Multi-Image GLaMM:} GLaMM \cite{rasheed2023glamm} is a strong single-image segmentation-based LVLM constructed for the Grounded Conversation Generation (GCG) task involving multi-turn pixel-grounded dialogue. GLaMM combines a Vicuna-based language backbone with SAM, similarly to LISA, and a novel RoIAlign-based region encoder. We implement multi-image processing for GLaMM by sequentially feeding the mask decoder with the encoded images and corresponding segmentation tokens to obtain segmentation masks. Both \modelname{} and Multi-Image GLaMM are initialized from GLaMM's pretrained model weights.\footnote{\scriptsize \url{https://huggingface.co/MBZUAI/GLaMM-GranD-Pretrained}.}
Since LISA and GLaMM are not natively designed to handle multiple images, we adapt \modelname{}'s implementation for distinguishing segmentation tokens belonging to different images. This is achieved by appending image identifiers to the \texttt{[SEG]} tokens, \eg, \texttt{(IMAGE1)}, enabling the models to process the image-specific token sets separately.
\end{itemize}

\subsection{Evaluation Metrics}
\label{appendix:eval-metrics}
Our evaluation consists of two parts: (1) segmentation quality and (2) semantic label accuracy for objects and parts. To evaluate performance on segmentation tasks, we employ mean Intersection over Union (\textbf{mIoU}), a widely used metric that measures the average overlap between predicted and ground truth masks across all classes. We also employ Average Precision at a 50\% IoU threshold (\textbf{AP50}), 
which measures the average precision across all recall levels at a fixed IoU threshold. Specifically, AP50 considers a predicted mask correct if its IoU with the ground truth mask is at least 50\%, summarizing model performance at this specific threshold. Additionally, following GLaMM \cite{rasheed2023glamm}, we report \textbf{Recall}, which evaluates region-specific grounding by using a two-phase validation approach based on IoU and SentenceBERT \cite{reimers2019sentence} similarity thresholds. In particular, a prediction is considered a true positive if its mask has IoU $\geq$ 50\% with the ground truth and its text label exceeds 50\% similarity with the ground truth.

To evaluate semantic labeling performance for the segmented objects and parts, we compute Semantic Similarity and Semantic IoU. Semantic Similarity (\textbf{SS}) measures similarity between predicted and true labels in the SentenceBERT \cite{reimers2019sentence} embedding space, following prior work \cite{yuan2023osprey, conti2024vocabulary}, while Semantic IoU (\textbf{S-IoU}) computes token-level overlap between predicted and ground truth labels, defined as
\begin{equation*}
\text{S-IoU} = \frac{1}{N} \sum_{i=1}^{N} \frac{|V(y_i) \cap V(\hat{y}_i)|}{|V(y_i) \cup V(\hat{y}_i)|},
\end{equation*}
where \( V(y) \) denotes the set of words comprising label \( y \).

\begin{table*}[t!]
\centering
\begin{tabular}{l|c|cccccc}
\toprule
\multirow{2}{*}{\textbf{Method}} & \multirow{2}{*}{Recall$\uparrow$} & \multirow{2}{*}{\# image tokens$\downarrow$} & \multirow{2}{*}{TFLOPS$\downarrow$} & \multicolumn{4}{c}{inference time$\downarrow$} \\
\cline{5-8}
& & & & CO & CP & UP & All \\
\hline \hline
Multi-Image GLaMM \cite{rasheed2023glamm} & 54.9 & 576 & 42.3 & 8.4 & 21.3 & 26.3 & 18.7 \\
Multi-Image LISA \cite{lai2024lisa} & 55.5 & 256 & 44.0 & 5.5 & 14.9 & 18.9& 13.1 \\
\hline
\cc \textbf{\modelname{}} & \cc \textbf{59.7} & \cc \textbf{32} & \cc \textbf{28.5} & \cc \textbf{4.3} & \cc \textbf{10.2} & \cc \textbf{12.3} & \cc \textbf{9.1} \\
\bottomrule
\end{tabular}
\vspace{-0.3cm}
\caption{\textbf{\modelname{} Efficiency in Numbers}. Alongside TFLOPS, we show inference time per sample in seconds for the common objects (CO), common parts (CP), unique parts (UP) tasks, as well as the average (All).}
\label{tab:appendix-efficiency}
\end{table*}

\section{Additional Experiments} \label{appendix:additional-experiments}
\subsection{Computational Efficiency Evaluation}
\label{appendix:efficiency}
For multi-image tasks, computational efficiency is critical to ensure scalability in inference and training as the number of images increases. To address this, \modelname{} employs Q-Former to query image embeddings, reducing the number of tokens per image to just 32, which is 8 times fewer tokens than LISA (256 tokens) and 18 times fewer than GLaMM (576 tokens). As shown in Table \ref{tab:appendix-efficiency}, this reduces TFLOPS by 35.23\%  over LISA and 32.62\% over GLaMM (LISA:44.0 \vs GLaMM:42.3 \vs \modelname{}:28.5 TFLOPS), yielding a $\sim$1.5$\times$speed-up and a 7.6\% relative performance gain over LISA.
These results demonstrate that \modelname{} offers substantial improvements in both computational efficiency and task performance compared to strong baselines.

\subsection{Per-Task Experimental Results}
Table \ref{tab:mixedparts-results-3tasks} presents results decomposed into the 3 \benchmarkname{} subtasks (common objects, common parts, and unique parts). 
The decreasing performance across all models delineates the incremental difficulty of the tasks, \ie, with common objects being the easiest task, followed by common parts, and finally unique parts as the most challenging task. \modelname{} consistently outperforms all baselines across all 3 tasks, with strong improvements in scores across all metrics compared to the next best baseline.

\begin{table*}[t!]
\centering
\begin{tabular}{l|l|ccc|cc}
\toprule
    \textbf{Task} & \textbf{Method} & AP50 & mIoU & Recall & SS & S-IoU \\ \hline \hline
    \multirow{6}{*}{\textbf{Common Objects}} & Cascade \cite{huang2023sparkles, achiam2023gpt, lai2024lisa} & 7.9 & 37.0 & 28.2 & 38.1 & 29.6 \\
    & Multi-Image PartGLEE \cite{li2024partglee} & 1.5 & 33.5 & 9.7 & 87.2 & 82.8 \\
    & Multi-Image VLPart \cite{sun2023going} & 18.2 & 42.4 & 45.6 & 58.8 & 55.2 \\
    & Multi-Image GLaMM \cite{rasheed2023glamm} & 63.2 & 71.6 & 73.3 & 86.5 & 86.2 \\
    & Multi-Image LISA \cite{lai2024lisa} & 60.2 & 70.0 & 71.9 & 86.0 & 85.3 \\
    & \cc \textbf{\modelname{}} & \cc \textbf{69.2} & \cc \textbf{75.2} & \cc \textbf{78.5} & \cc \textbf{93.7} & \cc \textbf{93.4} \\ \hline
    \multirow{6}{*}{\textbf{Common Parts}} & Cascade \cite{huang2023sparkles, achiam2023gpt, lai2024lisa} & 3.5 & 18.6 & 11.7 & 25.6 & 6.5 \\
    & Multi-Image PartGLEE \cite{li2024partglee} & 1.9 & 32.0 & 10.5 & \textbf{78.4} & 63.3 \\
    & Multi-Image VLPart \cite{sun2023going} & 15.7 & 44.5 & 29.1 & 54.4 & 39.9 \\
    & Multi-Image GLaMM \cite{rasheed2023glamm} & 36.7 & 52.7 & 48.4 & 68.5 & 59.0 \\
    & Multi-Image LISA \cite{lai2024lisa} & 37.3 & 54.2 & 50.5 & 73.9 & 63.4 \\
    & \cc \textbf{\modelname{}} & \cc \textbf{38.4} & \cc \textbf{56.9} & \cc \textbf{51.9} & \cc \underline{74.4} & \cc \textbf{64.4} \\ \hline
    \multirow{6}{*}{\textbf{Unique Parts}} & Cascade \cite{huang2023sparkles, achiam2023gpt, lai2024lisa} & 5.7 & 28.1 & 17.2 & 32.8 & 8.2 \\
    & Multi-Image PartGLEE \cite{li2024partglee} & 0.1 & 22.3 & 8.8 & 69.9 & 43.9 \\
    & Multi-Image VLPart \cite{sun2023going} & 6.4 & 41.5 & 29.0 & 64.0 & 44.4 \\
    & Multi-Image GLaMM \cite{rasheed2023glamm} & 28.9 & 55.3 & 43.1 & 75.3 & 68.4 \\
    & Multi-Image LISA \cite{lai2024lisa} & 26.6 & 55.0 & 44.1 & 76.2 & 68.7 \\
    & \cc \textbf{\modelname{}} & \cc \textbf{30.2} & \cc \textbf{59.1} & \cc \textbf{48.8} & \cc \textbf{80.1} & \cc \textbf{73.4} \\ \hline
    \multirow{6}{*}{\textbf{\benchmarkname{}}} & Cascade \cite{huang2023sparkles, achiam2023gpt, lai2024lisa} & 5.7 & 27.9 & 19.0 & 32.2 & 14.8 \\
    & Multi-Image PartGLEE \cite{li2024partglee} & 1.2 & 29.3 & 9.7 & 78.5 & 63.3 \\
    & Multi-Image VLPart \cite{sun2023going} & 13.4 & 42.8 & 34.6 & 59.1 & 46.5 \\
    & Multi-Image GLaMM \cite{rasheed2023glamm} & 42.9 & 59.9 & 54.9 & 76.8 & 71.2 \\
    & Multi-Image LISA \cite{lai2024lisa} & 41.4 & 59.7 & 55.5 & 78.7 & 72.5 \\
    & \cc \textbf{\modelname{}} & \cc \textbf{45.9} & \cc \textbf{63.7} & \cc \textbf{59.7} & \cc \textbf{82.7} & \cc \textbf{77.1} \\
    \bottomrule
\end{tabular}
\vspace{-0.2cm}
\caption{\textbf{Per-Task Experimental Results on \benchmarkname{}}. The first three metrics are segmentation-based while the last two are text-based. \modelname{} surpasses baselines across all three \benchmarkname{} tasks (common objects, common parts, and unique parts).}
\label{tab:mixedparts-results-3tasks}
\end{table*}

\begin{figure*}
    \centering
    \begin{tcolorbox}[
    enhanced,
    width=\textwidth,
    arc=3mm,
    boxsep=0mm,
    left=5mm,
    right=5mm,
    top=5mm,
    bottom=5mm,
    ]
    {\color{Mahogany}\textbf{Sparkles:}}
    \begin{itemize}
        \item Extracting all objects from both images:
        \begin{lstlisting}
What are the objects in IMAGE#1<Img><ImageHere></Img> and IMAGE#2<Img><ImageHere></Img>?
Reply with only the objects that visible in the images. Only talk about visible features, and limit output to the different objects. Make sure that each object is covered and described properly. Please talk about both the images and do not repeat objects.
\end{lstlisting}

        \item Extracting all parts from both images:

        \begin{lstlisting}
What are the parts of objects in IMAGE#1<Img><ImageHere></Img> and IMAGE#2<Img><ImageHere></Img>?
Reply with only the parts of objects that visible in the images. Only talk about visible features, and limit output to the different object and parts. Make sure that each part is covered and described properly. Please talk about both the images and do not repeat parts.
\end{lstlisting}
    \end{itemize}

    {\color{Mahogany}\textbf{GPT4o:}}
    \begin{itemize}
        \item Common objects:
        \begin{lstlisting}
You will be given description of two images. Understand the description of the images, make a list of the objects in both the images, and identify common objects. If you do not find any common objects, you can leave it empty.
Return these as a JSON object, strictly in the following format, output nothing else:
{
    'common_objects': ['object1', 'object2', ...]
}
replacing objectX with the actual object names. Or if there are no common objects:
{
    'common_objects': []
}
Here's the description of both the images: <@\textcolor{gray}{\textit{\textless{}Sparkles' outputs\textgreater{}}}@>.
Please maintain the JSON output format with object names for both images.
        \end{lstlisting}
        \item Common parts:
        \begin{lstlisting}
You will be given description of two images. Understand the description of the objects, make a list of the objects and their parts in both the images, and identify the common parts from all the objects in the images. If you do not find any common parts, you can leave it empty.
Return these as a JSON object, strictly in the following format, output nothing else:
{
    'common_parts': [
        {'image1': [object1, part1], 'image2': [object1, part1]}, 
        {'image1': [object1, part2], 'image2': [object1, part2]}, ...
        {'image1': [object2, part1], 'image2': [object2, part1]}, ...]
}
replacing objectX and partX with the actual object and part names. Or if there are no common parts:
{
    'common_parts': []
}
Here's the description of both the images: <@\textcolor{gray}{\textit{\textless{}Sparkles' outputs\textgreater{}}}@>.
Please maintain the JSON output format with both object and part names for both images.
        \end{lstlisting}
        \item Unique parts:
        \begin{lstlisting}
You will be given description of two images. Understand the description of the objects, make a list of the objects and their parts in both the images, and identify the unique parts from all the objects in the image. If you do not find any unique parts, you can leave it empty.
Return these as a JSON object, strictly in the following format, output nothing else:
{
    'unique_parts_image1': [
            ['object1','part1'], ['object1','part2'], ['object2','part1'], ...
        ]
    'unique_parts_image2': [
            ['object1','part1'], ['object1','part2'], ['object2','part1'], ...
        ]
}
replacing objectX and partX with the actual object and part names. Or if there are no unique parts:
{
    'unique_parts_image1': [],
    'unique_parts_image2': []
}
Here's the description of both the images: <@\textcolor{gray}{\textit{\textless{}Sparkles' outputs\textgreater{}}}@>.
Please maintain the JSON output format with both object and part name for both images.
        \end{lstlisting}
    \end{itemize}
    
    {\color{Mahogany}\textbf{LISA:}}
    \begin{itemize}
        \item Repeat for all common objects:
\begin{lstlisting}
Can you please segment <@\textcolor{gray}{\textit{\textless{}object\textgreater{}}}@> in this image?
\end{lstlisting}
        
        \item Repeat for all (common or unique) parts:
\begin{lstlisting}
Can you please segment <@\textcolor{gray}{\textit{\textless{}part\textgreater{}}}@> of <@\textcolor{gray}{\textit{\textless{}object\textgreater{}}}@> in this image?
\end{lstlisting}
    \end{itemize}
    \end{tcolorbox}
    \vspace{-0.4cm}
    \caption{\textbf{Prompts for the Cascaded Pipeline}.}
    \label{fig:cascade-prompts}
\end{figure*}
\begin{figure*}[t!]
  \centering
  \includegraphics[width=0.8\linewidth]{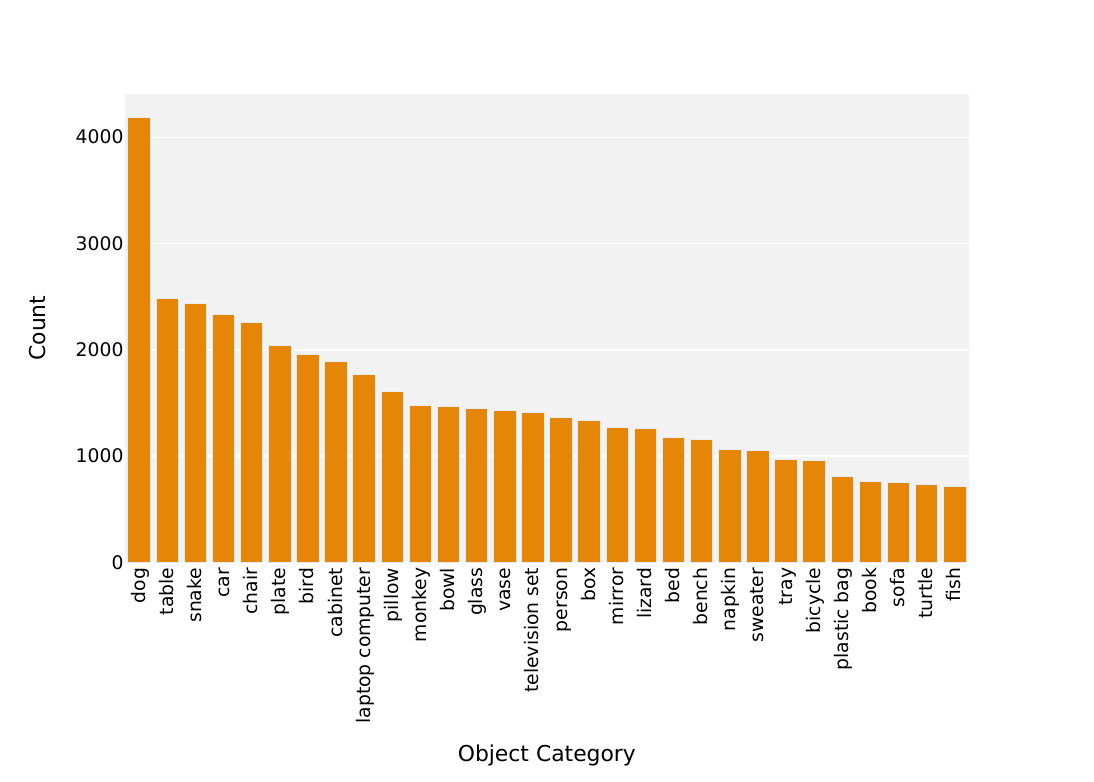}
  \caption{\textbf{Top-30 Most Frequent Object Categories in \benchmarkname{}.}}
  \label{fig:object-distribution}
\end{figure*}
\begin{figure*}[t!]
  \centering
  \includegraphics[width=0.8\linewidth]{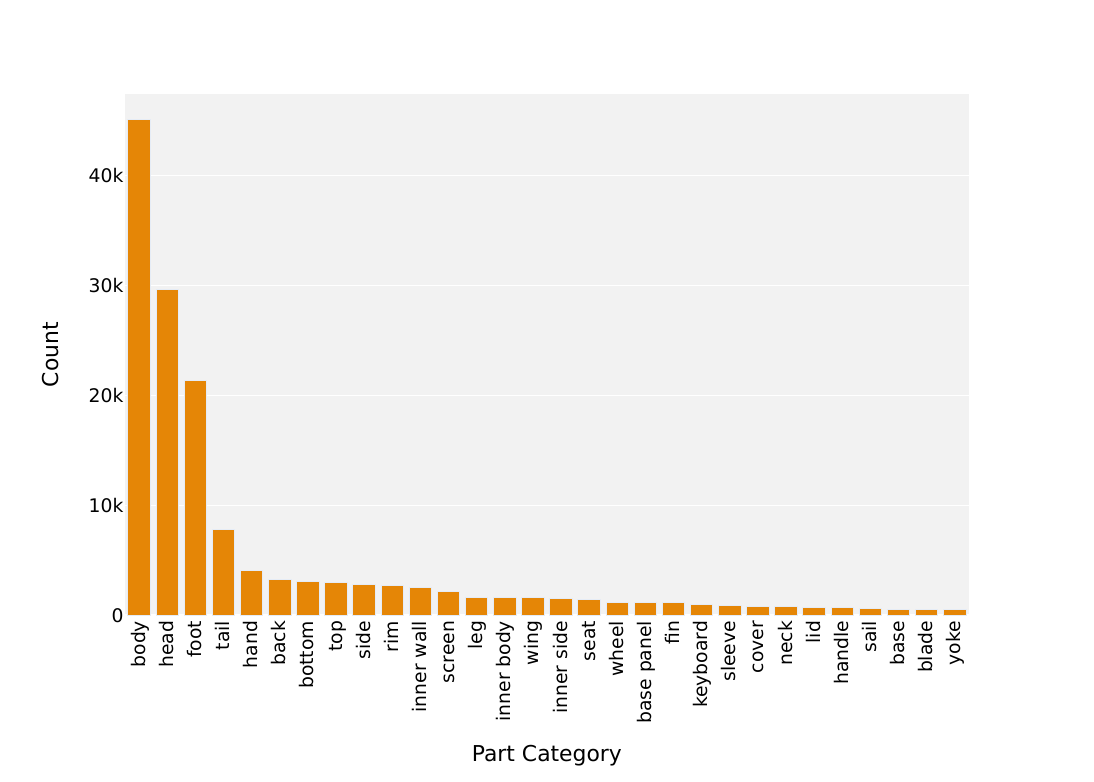}
  \caption{\textbf{Top-30 Most Frequent Part Categories in \benchmarkname{}.}}
  \label{fig:part-distribution}
  \vspace{-0.2cm}
\end{figure*}
\begin{figure*}[t!]
  \centering
  \includegraphics[width=0.8\linewidth]{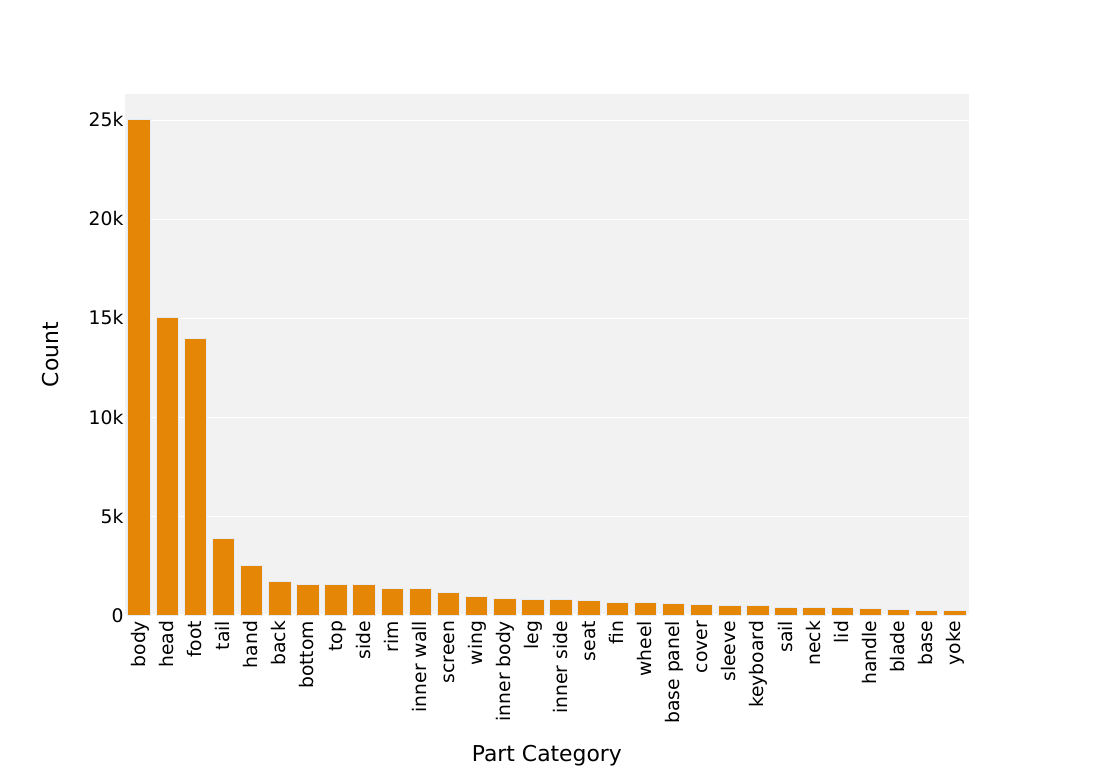}
  \caption{\textbf{Top-30 Most Frequent Common Parts in \benchmarkname{}.}}
  \label{fig:common-part-distribution}
\end{figure*}
\begin{figure*}[t!]
  \centering
  \includegraphics[width=0.8\linewidth]{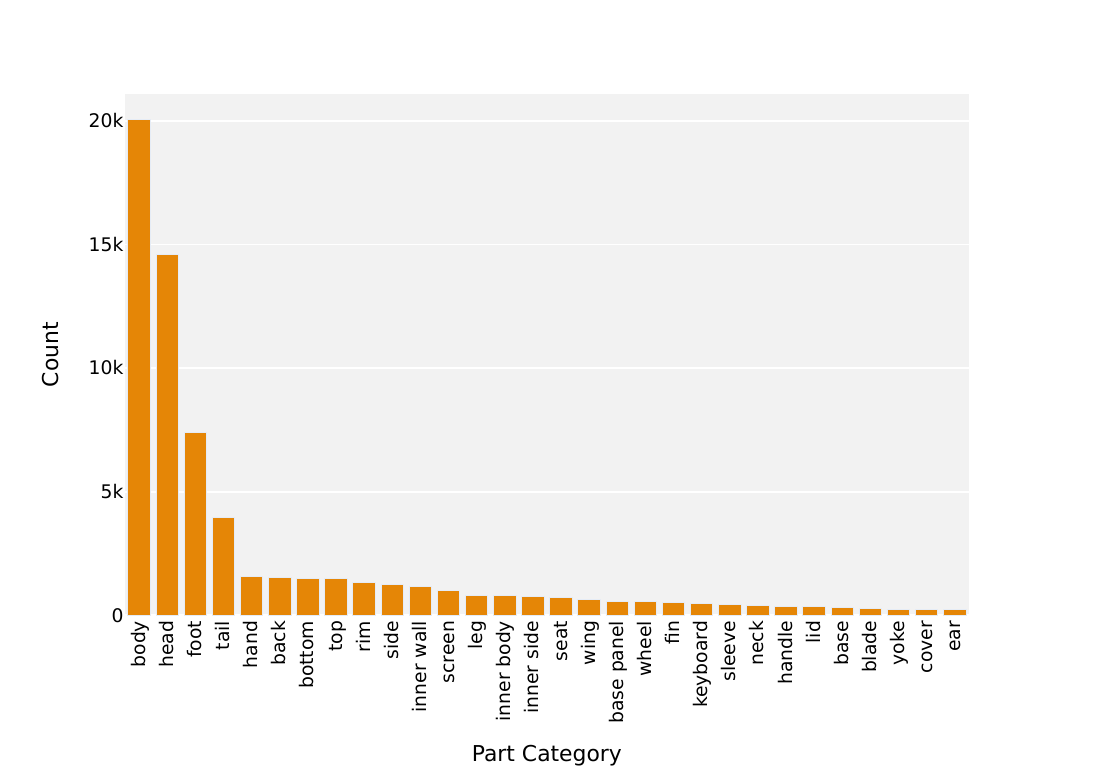}
  \caption{\textbf{Top-30 Most Frequent Unique Parts in \benchmarkname{}.}}
  \label{fig:unique-part-distribution}
\end{figure*}
\begin{table*}[t!]
\centering
\scriptsize
\resizebox{\linewidth}{!}{
\begin{tabulary}{\textwidth}{p{.12\textwidth}p{.12\textwidth}p{.8\textwidth}}
    \toprule
    \textbf{Dataset} & \textbf{Object} & \textbf{Parts} \\ \midrule
    ADE20K-Part234 & airplane & door, fuselage, landing gear, propeller, stabilizer, turbine engine, wing \\
    ADE20K-Part234 & armchair & apron, arm, back, back pillow, leg, seat, seat base \\
    ADE20K-Part234 & bed & footboard, headboard, leg, side rail \\
    ADE20K-Part234 & bench & arm, back, leg, seat \\
    ADE20K-Part234 & bookcase & door, drawer, front, side \\
    ADE20K-Part234 & bus & bumper, door, headlight, license plate, logo, mirror, wheel, window, wiper \\
    ADE20K-Part234 & cabinet & door, drawer, front, shelf, side, skirt, top \\
    ADE20K-Part234 & car & bumper, door, headlight, hood, license plate, logo, mirror, wheel, window, wiper \\
    ADE20K-Part234 & chair & apron, arm, back, base, leg, seat, skirt, stretcher \\
    ADE20K-Part234 & chandelier & arm, bulb, canopy, chain, cord, highlight, light source, shade \\
    ADE20K-Part234 & chest of drawers & apron, door, drawer, front, leg \\
    ADE20K-Part234 & clock & face, frame \\
    ADE20K-Part234 & coffee table & leg, top \\
    ADE20K-Part234 & computer & computer case, keyboard, monitor, mouse \\
    ADE20K-Part234 & cooking stove & burner, button panel, door, drawer, oven, stove \\
    ADE20K-Part234 & desk & apron, door, drawer, leg, shelf, top \\
    ADE20K-Part234 & dishwasher & button panel, handle, skirt \\
    ADE20K-Part234 & door & door frame, handle, knob, panel \\
    ADE20K-Part234 & fan & blade, canopy, tube \\
    ADE20K-Part234 & glass & base, bowl, opening, stem \\
    ADE20K-Part234 & kitchen island & door, drawer, front, side, top \\
    ADE20K-Part234 & lamp & arm, base, canopy, cord, highlight, light source, pipe, shade, tube \\
    ADE20K-Part234 & light & aperture, canopy, diffusor, highlight, light source, shade \\
    ADE20K-Part234 & microwave & button panel, door, front, side, top, window \\
    ADE20K-Part234 & minibike & license plate, mirror, seat, wheel \\
    ADE20K-Part234 & ottoman & back, leg, seat \\
    ADE20K-Part234 & oven & button panel, door, drawer, top \\
    ADE20K-Part234 & person & arm, back, foot, gaze, hand, head, leg, neck, torso \\
    ADE20K-Part234 & pool table & bed, leg, pocket \\
    ADE20K-Part234 & refrigerator & button panel, door, drawer, side \\
    ADE20K-Part234 & sconce & arm, backplate, highlight, light source, shade \\
    ADE20K-Part234 & shelf & door, drawer, front, shelf \\
    ADE20K-Part234 & sink & bowl, faucet, pedestal, tap, top \\
    ADE20K-Part234 & sofa & arm, back, back pillow, leg, seat base, seat cushion, skirt \\
    ADE20K-Part234 & stool & leg, seat \\
    ADE20K-Part234 & swivel chair & back, base, seat, wheel \\
    ADE20K-Part234 & table & apron, drawer, leg, shelf, top, wheel \\
    ADE20K-Part234 & television receiver & base, buttons, frame, keys, screen, speaker \\
    ADE20K-Part234 & toilet & bowl, cistern, lid \\
    ADE20K-Part234 & traffic light & housing, pole \\
    ADE20K-Part234 & truck & bumper, door, headlight, license plate, logo, mirror, wheel, window \\
    ADE20K-Part234 & van & bumper, door, headlight, license plate, logo, mirror, taillight, wheel, window, wiper \\
    ADE20K-Part234 & wardrobe & door, drawer, front, leg, mirror, top \\
    ADE20K-Part234 & washer & button panel, door, front, side \\
    PACO-LVIS & basket & base, bottom, cover, handle, inner side, rim, side \\
    PACO-LVIS & belt & bar, buckle, end tip, frame, hole, loop, prong, strap \\
    PACO-LVIS & bench & arm, back, leg, seat, stretcher, table top \\
    PACO-LVIS & bicycle & basket, down tube, fork, gear, handlebar, head tube, pedal, saddle, seat stay, seat tube, stem, top tube, wheel \\
    PACO-LVIS & blender & base, blade, cable, cover, cup, food cup, handle, inner body, seal ring, spout, switch, vapour cover \\
    PACO-LVIS & book & cover, page \\
    PACO-LVIS & bottle & base, body, bottom, cap, capsule, closure, handle, heel, inner body, label, neck, punt, ring, shoulder, sipper, spout, top \\
    PACO-LVIS & bowl & base, body, bottom, inner body, rim \\
    PACO-LVIS & box & bottom, inner side, lid, side \\
    PACO-LVIS & broom & brush, brush cap, handle, lower bristles, ring, shaft \\
    PACO-LVIS & bucket & base, body, bottom, cover, handle, inner body, loop, rim \\
    PACO-LVIS & calculator & body, key \\
    PACO-LVIS & can & base, body, bottom, inner body, lid, pull tab, rim, text \\
    PACO-LVIS & car & antenna, bumper, fender, grille, handle, headlight, hood, logo, mirror, rim, roof, runningboard, seat, sign, splashboard, steeringwheel, taillight, tank, trunk, turnsignal, wheel, window, windowpane, windshield, wiper \\
    PACO-LVIS & carton & bottom, cap, inner side, lid, side, tapering top, text, top \\
    PACO-LVIS & cellular telephone & back cover, bezel, button, screen \\
    PACO-LVIS & chair & apron, arm, back, base, leg, rail, seat, skirt, spindle, stile, stretcher, swivel, wheel \\
    PACO-LVIS & clock & base, cable, case, decoration, finial, hand, pediment \\
    PACO-LVIS & crate & bottom, handle, inner side, lid, side \\
    PACO-LVIS & cup & base, handle, inner body, rim \\
    PACO-LVIS & dog & body, ear, eye, foot, head, leg, neck, nose, tail, teeth \\
    PACO-LVIS & drill & body, handle \\
    PACO-LVIS & drum & base, body, cover, head, inner body, loop, lug, rim \\
    PACO-LVIS & earphone & cable, ear pads, headband, housing, slider \\
    PACO-LVIS & fan & base, blade, bracket, canopy, fan box, light, logo, motor, pedestal column, rod, string \\
    PACO-LVIS & glass & base, body, bottom, inner body, rim \\
    PACO-LVIS & guitar & back, body, bridge, fingerboard, headstock, hole, key, pickguard, side, string \\
    PACO-LVIS & hammer & face, grip, handle, head \\
    PACO-LVIS & handbag & base, body, bottom, handle, inner body, rim, zip \\
    PACO-LVIS & hat & inner side, logo, pom pom, rim, strap, visor \\
    PACO-LVIS & helmet & face shield, inner side, logo, rim, strap, visor \\
    PACO-LVIS & jar & base, body, bottom, cover, handle, inner body, lid, rim, sticker, text \\
    PACO-LVIS & kettle & base, body, cable, handle, inner body, lid, spout, switch \\
    PACO-LVIS & knife & blade, handle \\
    PACO-LVIS & ladder & foot, rail, step, top cap \\
    PACO-LVIS & lamp & base, bulb, cable, finial, pipe, shade, shade cap, shade inner side, switch \\
    PACO-LVIS & laptop computer & back, base panel, cable, camera, keyboard, logo, screen, touchpad \\
    PACO-LVIS & microwave oven & control panel, dial, door handle, inner side, side, time display, top, turntable \\
    \bottomrule
\end{tabulary}
}
\caption{\textbf{Object and Part Taxonomies by Dataset} (to be continued on next page).}
\label{tab:obj-parts1}
\vspace{-1em}
\end{table*}

\begin{table*}[t!]
\centering
\scriptsize
\resizebox{\linewidth}{!}{
\begin{tabulary}{\textwidth}{p{.1\textwidth}p{.15\textwidth}p{.74\textwidth}}
    \toprule
    \textbf{Dataset} & \textbf{Object} & \textbf{Parts} \\ \midrule
    PACO-LVIS & mirror & frame \\
    PACO-LVIS & mouse & body, left button, logo, right button, scroll wheel, side button, wire \\
    PACO-LVIS & mug & base, body, bottom, drawing, handle, inner body, rim, text \\
    PACO-LVIS & newspaper & text \\
    PACO-LVIS & pan & base, bottom, handle, inner side, lid, rim, side \\
    PACO-LVIS & pen & barrel, cap, clip, grip, tip \\
    PACO-LVIS & pencil & body, eraser, ferrule, lead \\
    PACO-LVIS & pillow & embroidery \\
    PACO-LVIS & pipe & colied tube, nozzle, nozzle stem \\
    PACO-LVIS & plastic bag & body, handle, hem, inner body, text \\
    PACO-LVIS & plate & base, body, bottom, inner wall, rim \\
    PACO-LVIS & pliers & blade, handle, jaw, joint \\
    PACO-LVIS & remote control & back, button, logo \\
    PACO-LVIS & scarf & body, fringes \\
    PACO-LVIS & scissors & blade, finger hole, handle, screw \\
    PACO-LVIS & screwdriver & handle, shank, tip \\
    PACO-LVIS & shoe & backstay, eyelet, heel, insole, lace, lining, outsole, quarter, throat, toe box, tongue, vamp, welt \\
    PACO-LVIS & slipper & insole, lining, outsole, strap, toe box, vamp \\
    PACO-LVIS & soap & base, body, bottom, cap, capsule, closure, handle, label, neck, punt, push pull cap, ring, shoulder, sipper, spout, top \\
    PACO-LVIS & sponge & rough surface \\
    PACO-LVIS & spoon & bowl, handle, neck, tip \\
    PACO-LVIS & stool & footrest, leg, seat, step \\
    PACO-LVIS & sweater & body, cuff, hem, neckband, shoulder, sleeve, yoke \\
    PACO-LVIS & table & apron, drawer, inner body, leg, rim, shelf, stretcher, top, wheel \\
    PACO-LVIS & tape & roll \\
    PACO-LVIS & telephone & back cover, bezel, button, screen \\
    PACO-LVIS & television set & base, bottom, button, side, top \\
    PACO-LVIS & tissue paper & roll \\
    PACO-LVIS & towel & body, border, hem, terry bar \\
    PACO-LVIS & trash can & body, bottom, hole, inner body, label, lid, pedal, rim, wheel \\
    PACO-LVIS & tray & base, bottom, inner side, inner wall, outer side, rim \\
    PACO-LVIS & vase & body, foot, handle, mouth, neck \\
    PACO-LVIS & wallet & flap, inner body \\
    PACO-LVIS & watch & buckle, case, dial, hand, lug, strap, window \\
    PACO-LVIS & wrench & handle, head \\
    PartImageNet & airplane (aeroplane) & body, engine, head, tail, wing \\
    PartImageNet & alligator (reptile) & body, foot, head, tail \\
    PartImageNet & antelope (quadruped) & body, foot, head, tail \\
    PartImageNet & ape (biped) & body, foot, hand, head, tail \\
    PartImageNet & badger (quadruped) & body, foot, head, tail \\
    PartImageNet & bear (quadruped) & body, foot, head, tail \\
    PartImageNet & bird (bird) & body, foot, head, tail, wing \\
    PartImageNet & boat (boat) & body, sail \\
    PartImageNet & camel (quadruped) & body, foot, head, tail \\
    PartImageNet & cat (quadruped) & body, foot, head, tail \\
    PartImageNet & cheetah (quadruped) & body, foot, head, tail \\
    PartImageNet & cougar (quadruped) & body, foot, head, tail \\
    PartImageNet & crocodile (reptile) & body, foot, head, tail \\
    PartImageNet & dog (quadruped) & body, foot, head, tail \\
    PartImageNet & fish (fish) & body, fin, head, tail \\
    PartImageNet & fox (quadruped) & body, foot, head, tail \\
    PartImageNet & frog (reptile) & body, foot, head, tail \\
    PartImageNet & goat (quadruped) & body, foot, head, tail \\
    PartImageNet & leopard (quadruped) & body, foot, head, tail \\
    PartImageNet & lizard (reptile) & body, foot, head, tail \\
    PartImageNet & mink (quadruped) & body, foot, head, tail \\
    PartImageNet & monkey (biped) & body, foot, hand, head, tail \\
    PartImageNet & otter (quadruped) & body, foot, head, tail \\
    PartImageNet & ox (quadruped) & body, foot, head, tail \\
    PartImageNet & panda (quadruped) & body, foot, head, tail \\
    PartImageNet & polecat (quadruped) & body, foot, head, tail \\
    PartImageNet & shark (fish) & body, fin, head, tail \\
    PartImageNet & sheep (quadruped) & body, foot, head, tail \\
    PartImageNet & snake (snake) & body, head \\
    PartImageNet & squirrel (quadruped) & body, foot, head, tail \\
    PartImageNet & swine (quadruped) & body, foot, head, tail \\
    PartImageNet & tiger (quadruped) & body, foot, head, tail \\
    PartImageNet & turtle (reptile) & body, foot, head, tail \\
    PartImageNet & water buffalo (quadruped) & body, foot, head, tail \\
    PartImageNet & weasel (quadruped) & body, foot, head, tail \\
    PartImageNet & whale (fish) & body, fin, head, tail \\
    PartImageNet & wolf (quadruped) & body, foot, head, tail \\
    \bottomrule
\end{tabulary}
}
\addtocounter{table}{-1} 
\caption{\textbf{Object and Part Taxonomies by Dataset} (continued).}
\label{tab:obj-parts2}
\vspace{-1em}
\end{table*}

\section{Limitations}
This research introduces \modelname{}, a model designed for part-focused semantic co-segmentation, incorporating several novel features. While our contributions include the proposal of a valuable new task and the release of a supporting dataset to encourage future research, there are a few limitations and assumptions that warrant discussion. First, \modelname{} assumes that semantically meaningful correspondences can be established between similar object parts across different images based on visual features. However, this assumption may break down in scenarios where visually similar parts serve different functions. Like many machine learning models, \modelname{} demonstrates promising results on curated datasets, but its generalizability to complex, real-world environments remains to be fully validated.
Future work should address these limitations, potentially through adaptive learning strategies that improve robustness and scalability across diverse, real-world applications.

\section{Broader Impact}
This work introduces several significant advancements with broad societal implications—both positive and cautionary.
The integration of object and part comparison methods holds promise for numerous applications in areas such as automated quality control in manufacturing, enhanced image-based search engines, and more sophisticated systems for digital content management and creation. 
For example, the proposed new task and models can automate and improve the accuracy of tasks that require detailed visual comparisons, such as quality assurance in manufacturing, potentially increasing efficiency while reducing human error. 
\modelname{} offers a more fine-grained understanding of images by identifying and segmenting objects across images based on their constituent common or unique parts. This can greatly aid in fields such as robotics, where agents require a detailed understanding of object parts for precise manipulation or obstacle avoidance tasks. Furthermore, in medical imaging, fine-grained visual decomposition can assist in detailed diagnostic tasks. Improved part-level understanding can also improve accessibility technologies, such as software for the visually impaired, by providing more descriptive and accurate visual summaries. 

However, reliance on visual segmentation alone may introduce risks in tasks where non-visual attributes are crucial. Objects with similar appearance but different material properties (\eg, plastic \vs metal) may be misinterpreted, potentially affecting tasks in high-stakes settings such as surgery or industrial handling of fragile materials. 
Depending on the application, the integration of other sensory data inputs (tactile, thermal, or acoustic sensors), together with visual data, would be beneficial in mitigating this risk. Future work can also design new datasets and models that consider metadata about material properties and other attributes, and ensure that the model is trained not just to recognize visual object and part similarities/differences, but also to associate these features with the correct properties. Finally, human-in-the-loop solutions that incorporate human oversight can ensure critical decisions are validated, improving reliability and trustworthiness in deployment.

\section{Image Attributions}

The three images in Figure \ref{tab:calico_in_context}, used for qualitative illustration of our model's in-context capabilities, are credited respectively to Erik Mclean, Pixabay, and Trace Constant on Pexels. All other images shown in this paper are sourced from publicly available datasets.

\end{document}